\documentclass[letterpaper, 10 pt, conference]{ieeeconf}  

\IEEEoverridecommandlockouts                              

\usepackage{hyperref}       
\usepackage{url}            
\usepackage{booktabs}       
\usepackage{amsfonts}       
\usepackage{nicefrac}       
\usepackage{microtype}      
\usepackage{xcolor}         
\usepackage{amsmath}
\usepackage[ruled]{algorithm2e}
\usepackage{multirow}
\usepackage{graphicx}
\usepackage{subfigure}
\usepackage{amssymb}
\usepackage{pifont}
\usepackage{makecell}
\let\labelindent\relax
\usepackage{enumitem}
\usepackage[symbol]{footmisc}
\usepackage{threeparttable}
\usepackage{bbding}
\usepackage{hyperref}
\hypersetup{hidelinks,
	colorlinks=true,
	allcolors=blue,
	pdfstartview=Fit,
	breaklinks=true}
	
\title{\LARGE \bf
Failure-aware Policy Learning for Self-assessable Robotics Tasks
}

\author{Kechun Xu, Runjian Chen, Shuqi Zhao, Zizhang Li, Hongxiang Yu, Ci Chen, Yue Wang, Rong Xiong
\thanks{This work was supported in part by the National Key R \& D Program of China under Grant 2021ZD0114500. Kechun Xu, Shuqi Zhao, Zizhang Li, Hongxiang Yu, Ci Chen, Yue Wang, Rong Xiong are with Zhejiang University, Hangzhou, China. Runjian Chen is with The University of Hong Kong. Corresponding author,{\tt\small wangyue@iipc.zju.edu.cn}.}%
}

\begin{document}

\maketitle
\thispagestyle{empty}
\pagestyle{empty}

\newcommand{\ba}{\mathbf{a}}\newcommand{\bA}{\mathbf{A}}
\newcommand{\bb}{\mathbf{b}}\newcommand{\bB}{\mathbf{B}}
\newcommand{\bc}{\mathbf{c}}\newcommand{\bC}{\mathbf{C}}
\newcommand{\bd}{\mathbf{d}}\newcommand{\bD}{\mathbf{D}}
\newcommand{\be}{\mathbf{e}}\newcommand{\bE}{\mathbf{E}}
\newcommand{\bff}{\mathbf{f}}\newcommand{\bF}{\mathbf{F}} %
\newcommand{\bg}{\mathbf{g}}\newcommand{\bG}{\mathbf{G}}
\newcommand{\bh}{\mathbf{h}}\newcommand{\bH}{\mathbf{H}}
\newcommand{\bi}{\mathbf{i}}\newcommand{\bI}{\mathbf{I}}
\newcommand{\bj}{\mathbf{j}}\newcommand{\bJ}{\mathbf{J}}
\newcommand{\bk}{\mathbf{k}}\newcommand{\bK}{\mathbf{K}}
\newcommand{\bl}{\mathbf{l}}\newcommand{\bL}{\mathbf{L}}
\newcommand{\bm}{\mathbf{m}}\newcommand{\bM}{\mathbf{M}}
\newcommand{\bn}{\mathbf{n}}\newcommand{\bN}{\mathbf{N}}
\newcommand{\bo}{\mathbf{o}}\newcommand{\bO}{\mathbf{O}}
\newcommand{\bp}{\mathbf{p}}\newcommand{\bP}{\mathbf{P}}
\newcommand{\bq}{\mathbf{q}}\newcommand{\bQ}{\mathbf{Q}}
\newcommand{\br}{\mathbf{r}}\newcommand{\bR}{\mathbf{R}}
\newcommand{\bs}{\mathbf{s}}\newcommand{\bS}{\mathbf{S}}
\newcommand{\bt}{\mathbf{t}}\newcommand{\bT}{\mathbf{T}}
\newcommand{\bu}{\mathbf{u}}\newcommand{\bU}{\mathbf{U}}
\newcommand{\bv}{\mathbf{v}}\newcommand{\bV}{\mathbf{V}}
\newcommand{\bw}{\mathbf{w}}\newcommand{\bW}{\mathbf{W}}
\newcommand{\bx}{\mathbf{x}}\newcommand{\bX}{\mathbf{X}}
\newcommand{\by}{\mathbf{y}}\newcommand{\bY}{\mathbf{Y}}
\newcommand{\bz}{\mathbf{z}}\newcommand{\bZ}{\mathbf{Z}}

\newcommand{\balpha}{\boldsymbol{\alpha}}\newcommand{\bAlpha}{\boldsymbol{\Alpha}}
\newcommand{\bbeta}{\boldsymbol{\beta}}\newcommand{\bBeta}{\boldsymbol{\Beta}}
\newcommand{\bgamma}{\boldsymbol{\gamma}}\newcommand{\bGamma}{\boldsymbol{\Gamma}}
\newcommand{\bdelta}{\boldsymbol{\delta}}\newcommand{\bDelta}{\boldsymbol{\Delta}}
\newcommand{\bepsilon}{\boldsymbol{\epsilon}}\newcommand{\bEpsilon}{\boldsymbol{\Epsilon}}
\newcommand{\bzeta}{\boldsymbol{\zeta}}\newcommand{\bZeta}{\boldsymbol{\Zeta}}
\newcommand{\beeta}{\boldsymbol{\eta}}\newcommand{\bEta}{\boldsymbol{\Eta}} %
\newcommand{\btheta}{\boldsymbol{\theta}}\newcommand{\bTheta}{\boldsymbol{\Theta}}
\newcommand{\biota}{\boldsymbol{\iota}}\newcommand{\bIota}{\boldsymbol{\Iota}}
\newcommand{\bkappa}{\boldsymbol{\kappa}}\newcommand{\bKappa}{\boldsymbol{\Kappa}}
\newcommand{\blambda}{\boldsymbol{\lambda}}\newcommand{\bLambda}{\boldsymbol{\Lambda}}
\newcommand{\bmu}{\boldsymbol{\mu}}\newcommand{\bMu}{\boldsymbol{\Mu}}
\newcommand{\bnu}{\boldsymbol{\nu}}\newcommand{\bNu}{\boldsymbol{\Nu}}
\newcommand{\bxi}{\boldsymbol{\xi}}\newcommand{\bXi}{\boldsymbol{\Xi}}
\newcommand{\bomikron}{\boldsymbol{\omikron}}\newcommand{\bOmikron}{\boldsymbol{\Omikron}}
\newcommand{\bpi}{\boldsymbol{\pi}}\newcommand{\bPi}{\boldsymbol{\Pi}}
\newcommand{\brho}{\boldsymbol{\rho}}\newcommand{\bRho}{\boldsymbol{\Rho}}
\newcommand{\bsigma}{\boldsymbol{\sigma}}\newcommand{\bSigma}{\boldsymbol{\Sigma}}
\newcommand{\btau}{\boldsymbol{\tau}}\newcommand{\bTau}{\boldsymbol{\Tau}}
\newcommand{\bypsilon}{\boldsymbol{\ypsilon}}\newcommand{\bYpsilon}{\boldsymbol{\Ypsilon}}
\newcommand{\bphi}{\boldsymbol{\phi}}\newcommand{\bPhi}{\boldsymbol{\Phi}}
\newcommand{\bchi}{\boldsymbol{\chi}}\newcommand{\bChi}{\boldsymbol{\Chi}}
\newcommand{\bpsi}{\boldsymbol{\psi}}\newcommand{\bPsi}{\boldsymbol{\Psi}}
\newcommand{\bomega}{\boldsymbol{\omega}}\newcommand{\bOmega}{\boldsymbol{\Omega}}

\newcommand{\nA}{\mathbb{A}}
\newcommand{\nB}{\mathbb{B}}
\newcommand{\nC}{\mathbb{C}}
\newcommand{\nD}{\mathbb{D}}
\newcommand{\nE}{\mathbb{E}}
\newcommand{\nF}{\mathbb{F}}
\newcommand{\nG}{\mathbb{G}}
\newcommand{\nH}{\mathbb{H}}
\newcommand{\nI}{\mathbb{I}}
\newcommand{\nJ}{\mathbb{J}}
\newcommand{\nK}{\mathbb{K}}
\newcommand{\nL}{\mathbb{L}}
\newcommand{\nM}{\mathbb{M}}
\newcommand{\nN}{\mathbb{N}}
\newcommand{\nO}{\mathbb{O}}
\newcommand{\nP}{\mathbb{P}}
\newcommand{\nQ}{\mathbb{Q}}
\newcommand{\nR}{\mathbb{R}}
\newcommand{\nS}{\mathbb{S}}
\newcommand{\nT}{\mathbb{T}}
\newcommand{\nU}{\mathbb{U}}
\newcommand{\nV}{\mathbb{V}}
\newcommand{\nW}{\mathbb{W}}
\newcommand{\nX}{\mathbb{X}}
\newcommand{\nY}{\mathbb{Y}}
\newcommand{\nZ}{\mathbb{Z}}

\newcommand{\cA}{\mathcal{A}}
\newcommand{\cB}{\mathcal{B}}
\newcommand{\cC}{\mathcal{C}}
\newcommand{\cD}{\mathcal{D}}
\newcommand{\cE}{\mathcal{E}}
\newcommand{\cF}{\mathcal{F}}
\newcommand{\cG}{\mathcal{G}}
\newcommand{\cH}{\mathcal{H}}
\newcommand{\cI}{\mathcal{I}}
\newcommand{\cJ}{\mathcal{J}}
\newcommand{\cK}{\mathcal{K}}
\newcommand{\cL}{\mathcal{L}}
\newcommand{\cM}{\mathcal{M}}
\newcommand{\cN}{\mathcal{N}}
\newcommand{\cO}{\mathcal{O}}
\newcommand{\cP}{\mathcal{P}}
\newcommand{\cQ}{\mathcal{Q}}
\newcommand{\cR}{\mathcal{R}}
\newcommand{\cS}{\mathcal{S}}
\newcommand{\cT}{\mathcal{T}}
\newcommand{\cU}{\mathcal{U}}
\newcommand{\cV}{\mathcal{V}}
\newcommand{\cW}{\mathcal{W}}
\newcommand{\cX}{\mathcal{X}}
\newcommand{\cY}{\mathcal{Y}}
\newcommand{\cZ}{\mathcal{Z}}

\newcommand{\figref}[1]{Fig.~\ref{#1}}
\newcommand{\secref}[1]{Section~\ref{#1}}
\newcommand{\algref}[1]{Algorithm~\ref{#1}}
\newcommand{\eqnref}[1]{Eq.~\eqref{#1}}
\newcommand{\tabref}[1]{Table~\ref{#1}}

\def\mc{\mathcal}
\def\mb{\mathbf}

\newcommand{\T}{^{\raisemath{-1pt}{\mathsf{T}}}}

\newcommand{\Perp}{\perp\!\!\! \perp}

\makeatletter
\DeclareRobustCommand\onedot{\futurelet\@let@token\@onedot}
\def\@onedot{\ifx\@let@token.\else.\null\fi\xspace}
\def\eg{e.g\onedot} \def\Eg{E.g\onedot}
\def\ie{i.e\onedot} \def\Ie{I.e\onedot}
\def\cf{cf\onedot} \def\Cf{Cf\onedot}
\def\etc{etc\onedot}
\def\vs{vs\onedot}
\def\wrt{wrt\onedot}
\def\dof{d.o.f\onedot}
\def\etal{et~al\onedot}
\def\iid{i.i.d\onedot}
\makeatother

\renewcommand\UrlFont{\color{blue}\rmfamily}

\newcommand*\rot{\rotatebox{90}}

\newcommand{\boldparagraph}[1]{\vspace{0.2cm}\noindent{\bf #1:} }

\definecolor{darkgreen}{rgb}{0,0.7,0}
\definecolor{lightred}{rgb}{1.,0.5,0.5}

\newcommand{\red}[1]{\noindent{\color{red}{#1}}}
\newcommand{\lightred}[1]{\noindent{\color{lightred}{#1}}}
\newcommand{\ag}[1]{ \noindent {\color{blue} {\bf Andreas:} {#1}} }
\newcommand{\lars}[1]{ \noindent {\color{cyan} {\bf Lars:} {#1}} }
\newcommand{\michael}[1]{ \noindent {\color{blue} {\bf Michael:} {#1}} }
\newcommand{\songyou}[1]{ \noindent {\color{red} {\bf Songyou:} {#1}} }

\begin{abstract}

    Self-assessment rules play an essential role in safe and effective real-world robotic applications, which verify the feasibility of the selected action before actual execution. But how to utilize the self-assessment results to re-choose actions remains a challenge. Previous methods eliminate the selected action evaluated as failed by the self-assessment rules, and re-choose one with the next-highest affordance~({\it i.e.} process-of-elimination strategy~\cite{ghosh2021generalization}), which ignores the dependency between the self-assessment results and the remaining untried actions. However, this dependency is important since the previous failures might help trim the remaining over-estimated actions. In this paper, we set to investigate this dependency by learning a failure-aware policy. We propose two architectures for the failure-aware policy by representing the self-assessment results of previous failures as the variable state, and leveraging recurrent neural networks to implicitly memorize the previous failures. Experiments conducted on three tasks demonstrate that our method can achieve better performances with higher task success rates by less trials. Moreover, when the actions are correlated, learning a failure-aware policy can achieve better performance than the process-of-elimination strategy.

\end{abstract}

\section{Introduction}

A crucial problem for robotic applications in real world is how to promise action safety and effectiveness, especially for the applications of robot learning policies in unseen testing scenarios. A common way to deal with this problem is to utilize pre-defined self-assessment rules, which verify the feasibility of the selected actions before actual robot execution. For example, most of works in autonomous driving~\cite{isele2018safe,srinivasan2020learning,krasowski2020safe,mokhtari2021safe,lin2018efficient,chen2021multi} utilize the pre-built global map to forecast potential collision with the selected action. Once the selected action fails to pass the self-assessment tests, the learned policy has to re-choose an action. However, since there is no actual action execution, the observation stays invariant. Thus, the same action will be re-chosen by the learned policy and fail again, which raises a problem: how to re-choose actions among the remaining untried ones? 

\begin{figure}[t]
  \centering
  \includegraphics[width=0.8\linewidth]{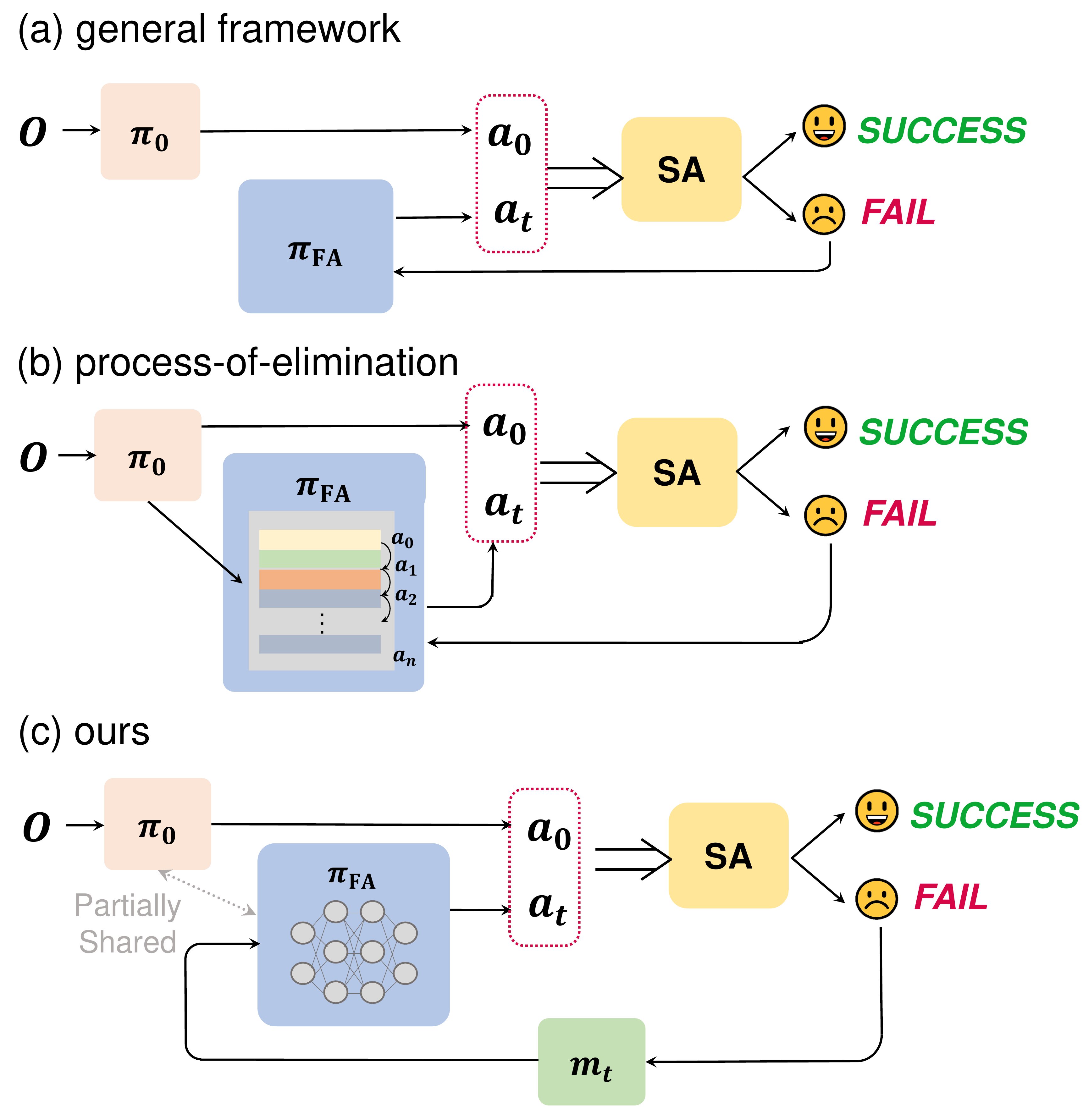}
  \vspace{-0.3cm}
  \caption{(a)~The general framework to utilize the self-assessment module. A learned policy $\pi_{0}$ takes as input the observation $o$, and selects the initial action $a_0$. Then a self-assessment module $\mb{SA}$ evaluates the selected action. If failed, a failure-aware policy $\pi_{\mathrm{FA}}$ will re-choose the action $a_t$ until getting successful feedback. (b)~A common pipeline (process-of-elimination~\cite{ghosh2021generalization}) to design $\pi_{\mathrm{FA}}$, which re-chooses action $a_t$ in a sorting way using the action affordance map generated from $\pi_{0}$. (c)~Our pipeline, which constructs a failure memory representation $m_t$ and uses a learning-based $\pi_{\mathrm{FA}}$.}
  \label{overview}
  \vspace{-0.7cm}
\end{figure}

Considering that the self-assessment results enable the re-decision of a sequence of actions, in this paper, we formally state this re-choosing process as a sequential decision making problem under an invariant observation. Specifically, as shown in Fig.~\ref{overview}(a), given an observation $o$, a learned policy $\pi_{0}(o)$ generates the initial action $a_0$, followed by a self-assessment module $\mb{SA}$ to indicate success or failure. If failed, the failure-aware policy $\pi_{\mathrm{FA}}$ is activated for re-choosing the action. Conditioned on the observation $o$, $\pi_{\mathrm{FA}}(s_t|o)$ takes the self-assessment results of the previous actions as state $s_t$ to generate a new action $a_t$ at step $t$ until getting successful feedback from $\mb{SA}$. Given the action affordance distribution predicted from $\pi_{0}$, one intuitive way to design $\pi_{\mathrm{FA}}$, which is shown in Fig.~\ref{overview}(b), is to choose the action with the next-highest affordance if the previously selected action is evaluated as unqualified, and so forth ({\it i.e.} process-of-elimination strategy~\cite{ghosh2021generalization,isele2018safe,srinivasan2020learning,krasowski2020safe,mokhtari2021safe,lin2018efficient,chen2021multi}). 
However, there raises a further question of whether it is an optimal failure-aware policy. In other words, {\it under invariant observation, does the equality hold between the action with the second-highest affordance and the action with the highest affordance conditioned on previous failures?} The process-of-elimination strategy gives a positive answer to this question, which means previous action failures cause no influence on the affordance distribution of the remaining untried actions. However, we argue that the previous failure is an important prior for the action re-choosing. Therefore, the equality might not always hold.

In this paper, we set to investigate the dependency between the self-assessment results and the remaining untried actions by learning the failure-aware policy $\pi_{\mathrm{FA}}$ (Fig.~\ref{overview}(c)). We define self-assessable robotics tasks as those where the robot can evaluate itself by some self-assessment rules before actual action execution. Our key insight is to integrate the self-assessment results during the observation-invariant process into the training of $\pi_{\mathrm{FA}}$. We represent the results of the previous failure verified by the self-assessment module as $m_t$, which serves as the variable state $s_t$ of $\pi_{\mathrm{FA}}$. Also, Recurrent Neural Networks~\cite{hochreiter1997long,cho2014learning} are helpful for the implicit representation to memorize the previous failures. Based on these points, we propose two architectures for the failure-aware policy. One tends to explicitly degenerate actions similar to failed ones, while another uses recurrent network to implicitly represent failure memory of the trial sequence. Experiments conducted on three tasks demonstrate that our method can achieve better performances with higher task success rates by less trials. Moreover, we find that when the actions are correlated, learning a failure-aware policy can achieve better performance than the process-of-elimination strategy. To summarize, our contributions are as follows:

\begin{itemize}[leftmargin=15pt, partopsep=0pt, topsep=0pt]
    \item Our main contribution is to provide a learning-based perspective to utilize self-assessment results to learn a failure-aware policy for self-assessable robotics tasks.
    \item We propose two effective architectures for the failure-aware policy. One tends to degenerate actions similar to failed ones, while another uses recurrent network to implicitly represent failure memory of the trial sequence.
    \item We evaluate our method with three typical self-assessable robotics tasks, including sequential image classification, object reorientation and localization. Both simulated and real-world experiments validate the effectiveness of our policy, and the two architectures present different advantages according to the task properties. Moreover, when the actions are correlated, learning a failure-aware policy can achieve better performance than the process-of-elimination strategy.
\end{itemize}

\section{Related Works}

\label{sec:related-work}

{\bf Robotic Self-assessment.} Recently, robotic self-assessment has become a topic of interest in human-robot interaction. \cite{burghouts2020robotic} highlights the importance of online {Competence Assessment (CA)} for safe real-world operation of robots. \cite{nortonmetrics} further extends the term to {Proficiency Self-Assessment (PSA)}, which shows the ability of a robot to predict, estimate, or measure its performance given a context or environment before action execution. Actually, this term can be extended to all robotics tasks, and a self-assessable robotics task means that the robot has some PSA metrics to evaluate its performance. For example, lots of works discard unsafe selected actions with prior knowledge of the global environment~\cite{isele2018safe,srinivasan2020learning,krasowski2020safe,mokhtari2021safe,lin2018efficient,chen2021multi}, or with estimated environment dynamics~\cite{deptula2019approximate,xu2021efficient,kumar2021error}, or by pre-acting with visual imagination~\cite{finn2017deep,ebert2018visual,wang2019learning,di2020safari}. These works either simply use self-assessment metrics to filter actions and re-choose the action with the next-highest affordance or handcrafted safer policy, or consume a large amount of data to build the environment or imagination module, which brings another problem of estimation bias. In contrast, our work directly uses the result representation from self-assessment during the training process, thus easily integrating self-assessment into our policy distribution. 

{\bf Failure-aware Policy Learning.} In this paper, we define the failure-aware policy to be aware of the previous action failures, and utilize the failed trials to predict more reliable actions. However, there are few studies under this definition. Hence, we review works that study policies predicting the success probability of the current action~\cite{xu2021efficient,finn2017deep,ebert2018visual,wang2019learning,di2020safari,xu2021push,xu2021learning}, or estimating the novelty and uncertainty of current observation \cite{richter2017safe, wellhausen2020safe, kahn2017uncertainty, lutjens2019safe} which can be referred to as studies of failure-prediction policy. Other works like~\cite{kumar2021error} predict the error of current action execution and propose an error-aware policy which takes as input the predicted future state error, and generates the corrected action. Similarly, \cite{wong2022error} conducts an error detector by checking the reconstruction of the current state. However, most of these policies only measure the success probability of the current actions, without awareness of previous failed trials.

{\bf Robotic Exploration.} Robotic exploration is a more general domain of our work, which can be regarded as exploration conditioned on the previous failures. Traditional works~\cite{martinez2007active,peters2010relative,sutton2018reinforcement} design algorithms to explore states with less visiting times ({\it i.e} with larger entropy). Also, many recent works follow the similar idea to set bonus to states deemed to be interesting or novel~\cite{houthooft2016vime,bellemare2016unifying,pathak2017curiosity,tang2017exploration,burda2018exploration,machado2020count}. Other works like~\cite{ghosh2021generalization} propose to train a set of policies to overlap a group of contexts with a disagreement penalty. Another view is to decouple the exploration policy from the exploitation policy, thus eliminating the inductive bias from task reward and stabilizing the policy training~\cite{whitney2021decoupled}.
\vspace{-0.3cm}

\begin{figure*}[t]
  \centering
  \includegraphics[width=0.86\linewidth]{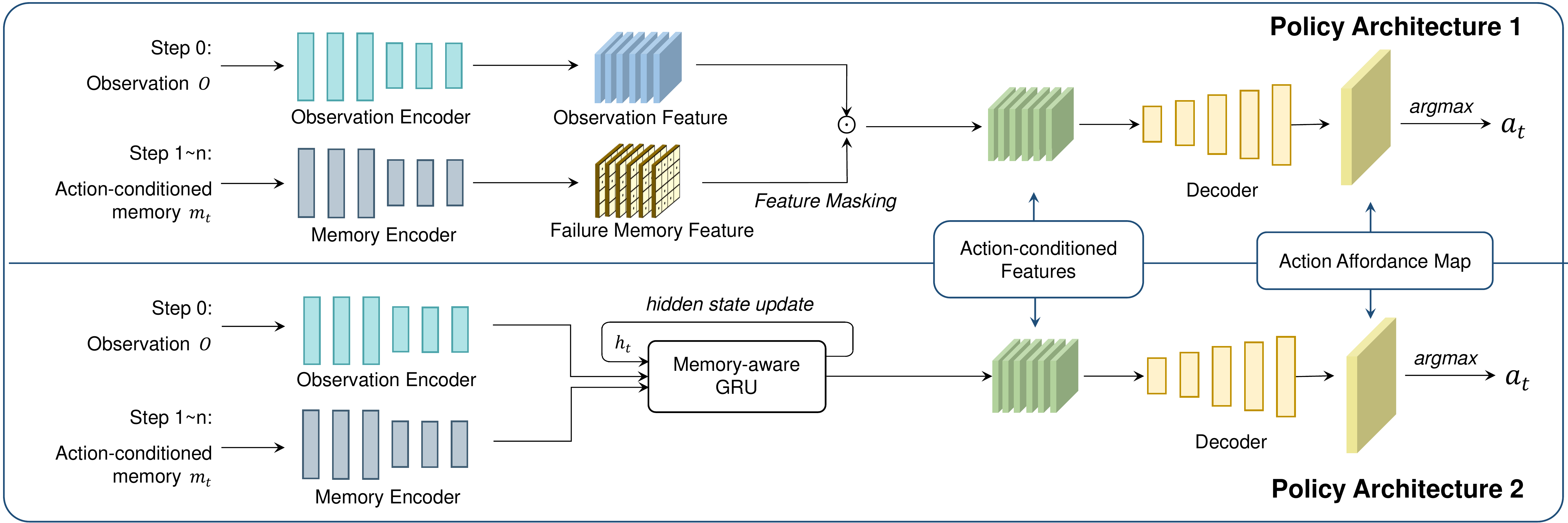}
  \vspace{-0.2cm}
  \caption{Two architectures for the failure-aware policy $\pi_{\mathrm{FA}}$. Note that the observation encoder and the decoder are components of the learned policy $\pi_{0}$, which accelerates the failure-aware policy training process.}
  \label{architecture}
  \vspace{-0.6cm}
\end{figure*}

\section{Problem Formulation}

\label{sec:formulation}

In this work, we define self-assessable robotics tasks as those where a robot has some self-assessment rules to evaluate its performance before actually executing actions, and the evaluation results enable action re-choosing. Such self-assessment rules often serve as a safe module in real applications by predicting collision with simulation, which can provide a relatively accurate failure awareness. As a result, in this paper, we assume that self-assessment rules that correctly distinguish failure. Since there is no action execution, the observation stays invariant. Following the general framework in Fig.~\ref{overview}(a), in this section, we formulate and compare the process-of-elimination strategy and our method. Given the invariant observation $o$, a discrete action set $\cA$, and a self-assessment module $\mb{SA}$, we can formulate the same part of the process-of-elimination strategy and our method as follows:
\begin{equation}
\begin{gathered}
a_{0}=\left.\pi_{0}(o)\right|_{a_{0} \in \cA} \\
f_{t}(a_t|o):=\left.\mb{SA}\left(o, a_{t}\right)\right|_{t \geq 0}=\left\{\begin{array}{cc}
1, & \text { if successful } \\
0, & \text { otherwise }
\end{array}\right. \\
a_{t+1}=\pi_{\mathrm{FA}}\left(f_{t}\right| o)|_{a_{t+1} \in \cA} \\
\end{gathered}
\end{equation}
where $\pi_{0}$ is the learned policy, $a_t$ is the selected action at step $t$, $f_t$ is a binary distribution defined by the $\mb{SA}$ results up to step $t$, and $\pi_{\mathrm{FA}}$ is a failure-aware policy conditioned on invariant observation $o$. Note that in this work, $\pi_{0}$ is assumed as a differentiable policy that contains an observation encoder and a decoder predicting the action affordances. 

\textbf{Process-of-elimination.} This is an intuitive way to design $\pi_{\mathrm{FA}}$ (Fig.~\ref{overview}(b)) which chooses the action with the next-highest affordance after figuring out that the selected action is unreliable, and can be formulated as follows:
\begin{equation}
a_{t+1}=\pi_{\mathrm{FA}}\left(f_{t}\right| o)|_{a_{t+1} \in \cA}=\pi_{0}(o) * f_{t}(a_t|o)|_{a_{t+1} \in \cA}
\end{equation}
In this formulation, $\pi_{\mathrm{FA}}$ is a handcrafted sorting policy, which simply multiplies the learned policy distribution by $f_t$. In this way, the previous action failures do not influence the affordance distribution of the remaining untried actions.

\textbf{Failure-aware Learning Policy.} In this paper, we propose to integrate the self-assessment results into the failure-aware policy training, and use the result representation $m_t$ from $f_t$ as the variable state (Fig.~\ref{overview}(c)). Our framework can be represented as follows:
\begin{equation}
\begin{gathered}
{m_{t}} \sim f_{t}(a_t|o)\\
a_{t+1}=\pi_{\mathrm{FA}}\left(m_{t}\right| o; \theta)|_{a_{t+1} \in \cA}
\end{gathered}
\end{equation}
where $\pi_{\mathrm{FA}}$ is a learnable failure-aware policy, $m_t$ is a representation of $f_t$ with the same size as the action set $\cA$, and $\theta$ represents the learnable parameters of neural network.

\textbf{Self-assessment Representation.} In our paper, $m_t$ is used to represent the results of self-assessment. Concretely, it is a binary or normalized matrix of which each element represents the trial memory of the corresponding action, thus named action-conditioned memory. $m_t$ is initialized at the beginning of every episode. For a binary $m_t$, it is initialized as an all-one matrix with the same size as the action set $\cA$, while for a normalized representation, it is initialized as the normalized affordance map predicted by the learned policy $\pi_{0}$. If an action is assessed as failed during the trial process, then the corresponding element is set to zero, thus updating $m_t$ during the whole episode.

\section{Methods}
\label{sec:method}
\subsection{System Overview}
Fig.~\ref{overview}(c) shows our pipeline. At the first step, the learned policy $\pi_{0}$ takes as input the observation $o$ to get an initial action $a_0$. If $a_0$ fails according to the self-assessment module $\mb{SA}$, a self-assessment result representation $m_t$ will be constructed and fed into the failure-aware policy $\pi_{\mathrm{FA}}$ to re-choose another action $a_t$. Note that the network parameters of $\pi_{0}$ and $\pi_{\mathrm{FA}}$ are partially shared. Compared to the process-of-elimination strategy, we propose to investigate the dependency between the self-assessment results and the remaining untried actions by learning the failure-aware policy.  

\subsection{Failure-aware Policy Architecture}
Considering that the memory of previous failure can be utilized either explicitly or implicitly, we propose two architectures for the failure-aware policy $\pi_{\mathrm{FA}}$, which are shown in Fig. \ref{architecture}. Both of them contain an observation encoder and a decoder, which are the shared components of the learned policy $\pi_{0}$, which accelerates the training process. 

\textbf{Policy Architecture 1.} The first proposed architecture is to encode the self-assessment result representation $m_t$ into the same shape of the embedding feature generated from the observation encoder, and conduct an element-wise product to mask the observation feature by the failure memory feature embedding, which generates an action-conditioned feature. In this way, failure memory is explicitly considered into the feature embedding with a mask-like operation, thus affecting a shift on the action distribution. 
\begin{equation}
\label{eq-1}
\begin{gathered}
e = \mb{E}_{\mathrm{o}}(o) \odot \mb{E}_{\mathrm{m}}(m_t) \\
a_{t}=\mathop{\mathrm{argmax}}\limits_{a_t \in \cA} \mb{D}(e)
\end{gathered}
\end{equation}
where $\mb{E}_{\mathrm{o}}$, $\mb{E}_{\mathrm{m}}$ and $\mb{D}$ symbolize the observation encoder, the memory encoder and the decoder respectively. $a_t$ is selected from the action affordance map generated from $\mb{D}$. Note that the memory encoder can be simplified as a replica transform to the shape of the observation feature, or an identity transform in specific implementations.

\textbf{Policy Architecture 2.} The second architecture aims to bring the failure memory as a recurrent form across the episode using a memory-aware module. In this architecture, the feature embedding comes from the observation $o$ at the first step, and from the updated $m_t$ in the following steps. In this way, the failure memory is implicitly delivered across the decision process as a latent embedding.
\begin{equation}
    e = 
    \begin{cases}
    \mb{E}_{\mathrm{o}}(o) & t=0 \\
    \mb{E}_{\mathrm{m}}(m_t) & t>0
    \end{cases}
\end{equation}
where $\mb{E}_{\mathrm{o}}$ and $\mb{E}_{\mathrm{m}}$ are of the same definitions in Eq.~\ref{eq-1}. With the feature embedding, the memory-aware module obtains awareness of the observation at the first step and implicitly represents it by the hidden vector, then produces new action distributions with recurrent memory in the following steps.
\begin{equation}
\begin{aligned}
a_{t}=& \mathop{\mathrm{argmax}}\limits_{a_t\in \cA} \mb{D}\left(\mb{GRU}(e)\right)
\end{aligned}
\end{equation}
where $\mb{D}$ is of the same definition in Eq.~\ref{eq-1}, and $\mb{GRU}$ corresponds to the memory-aware module.

\subsection{Policy Learning}

We implement behavior cloning to train the learned policy $\pi_{0}$. Note that there is no sequential decision making for this stage. For each step, the policy generates an action under the observation and gets feedback from self-assessment. Also, since parameters of $\pi_{0}$ and $\pi_{\mathrm{FA}}$ are partially shared, it can be seen as pre-training for the failure-aware policy $\pi_{\mathrm{FA}}$. 

To train the failure-aware policy, we apply value-based RL algorithms. For each episode, the policy is provided with the observation at the first step and chooses an action. If the feedback from the self-assessment module $\mb{SA}$ is positive, then the episode ends. Otherwise, the failure-aware policy $\pi_{\mathrm{FA}}$ will take as input the memory representation $m_t$ and re-choose an action until receiving positive feedback from $\mb{SA}$. During the training process, network parameters shared with $\pi_{0}$ will be fixed.

\section{Experimental Results}

\label{sec:evaluation}
In this section, we will conduct experiments in three self-assessable robotics tasks to: 1) evaluate the effectiveness and advantages of our failure-aware policy compared with other methods; 2) show the different performances of the two failure-aware policy architectures; 3) investigate what kind of policy is optimal for the sequential decision making problem under invariant observation.

\subsection{Experimental Setup}

\begin{figure}[t]
  \centering
  \includegraphics[width=0.88\linewidth]{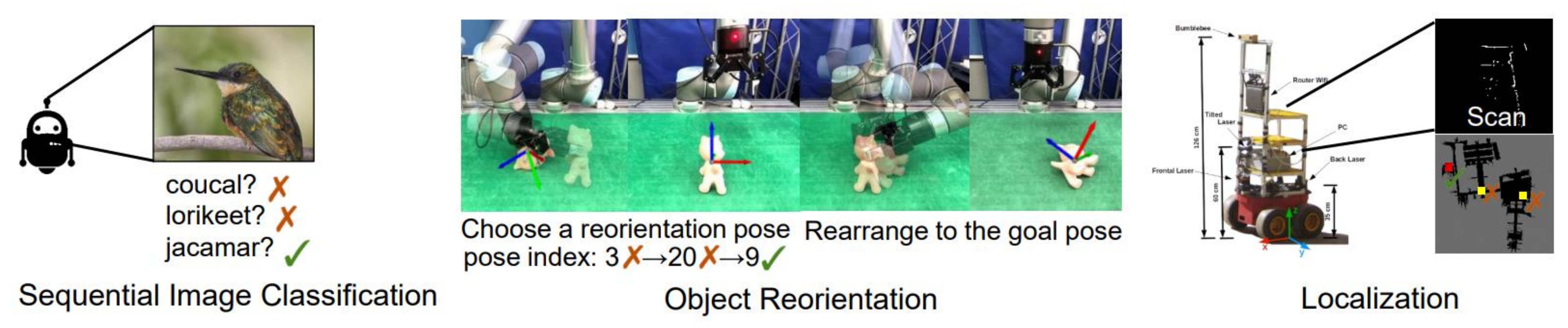}
  \vspace{-0.2cm}
  \caption{Three tasks for evaluation.} 
  \label{fig:task}
  \vspace{-0.1cm}
\end{figure}

We consider three typical self-assessable robotics tasks~(Fig.~\ref{fig:task}) for evaluation. The first task is sequential image classification on ImageNet~\cite{deng2009imagenet} motivated by \cite{ghosh2021generalization}. In this task, the robot observes an image at the beginning of an episode, and identifies a label for this image. After choosing a label, the self-assessment module will indicate whether the choice is correct or not. The second task is object reorientation, where a robot is supposed to choose a reorientation object pose to achieve a feasible pick-reorient-place process ({\it i.e.} successful path planning of the whole manipulation) with path planning cost as less as possible~\cite{xu2021efficient}. For this task, the policy is trained in SAPIEN~\cite{xiang2020sapien} with a UR5 arm, tested with unseen samples, and evaluated in real world. Self-assessment is conducted by path planning algorithms, which guarantee the execution success. And the last is localization on synthetic dataset \cite{chen2021deep} and real-world dataset UPO~\cite{ramon2014navigating} and Bicocca~\cite{bonarini2006rawseeds,ceriani2009rawseeds}, which predicts the position of the robot given a global map and an observed scan, and gets the assessment of the localization accuracy. In this task, if the predicted position is at the $k\times k$ neighborhood of the ground truth position, then the action is regarded as successful. Note that the robot will re-choose action after getting the self-assessment results until evaluated as successful or up to the limited trial times. In real applications, we can use registration algorithms
as the self-assessment module which measures localization accuracy. Details of the self-assessment module of these tasks, the learned policy, and implementations of our two architectures as well as the training and testing settings can be found in Appendix.

\subsection{Metrics and Baselines}

In this work, we aim to achieve a reliable action as soon as possible, since online self-assessment costs time and energy. Thus, we limit the sequential trial number to $t=5$ times, and measure the algorithms with the following metrics.
\begin{itemize}[leftmargin=10pt]
    \item \textbf{Task Success Rate (tsr)}: Average task success rate across all testing samples. If the policy passes the self-assessment within 5 trials, then the corresponding testing sample is regarded as successful.
    \item \textbf{Trial Number to Success (tns)}: Average trial number to get a success feedback from self-assessment of all success samples. 
\end{itemize}

All the tasks are measured with {\bf tsr} and {\bf tns}, which demonstrate the effectiveness and efficiency of action re-choosing based on previous failed trials. Also, an additional metric is tested for the object reorientation task:
\begin{itemize}[leftmargin=10pt]
    \item \textbf{Planning Cost (pc)}: Average path planning cost across all testing samples. It is a unique metric for object reorientation task, where robot is supposed to choose a reorientation pose with path planning cost as less as possible.
\end{itemize}

We compare the performance of our system to the following baseline approaches:

\textbf{Random Exploration (RE).} A policy which selects actions uniformly at random from the candidate action set $\cA$. 

\textbf{Learned Policy with Random Exploration (LPRE).} A policy that uses the learned policy $\pi_{0}$ for the first step, and if failed, then use the \textbf{RE} policy among the remaining actions.

\textbf{Sorting Policy (SP).} A policy that uses the process-of-elimination strategy.

Also, we name our policies of two architecture as  \textbf{Failure-aware Feature Masking Policy (FMP-1)} and \textbf{Failure-aware Recurrent Memory Policy (FMP-2)} respectively.

\begin{table}[t]
\caption{{Testing Performance of Sequential Image Classification}}
\vspace{-0.3cm}
\label{table:1}
\begin{center}
\begin{tabular}{p{1.8cm}|p{1.5cm}|p{1.5cm}}
    \hline
    \makecell[c]{Method} & \makecell[c]{tsr/\%} & \makecell[c]{tns} \\
    \hline
    \makecell[c]{RE} & \makecell[c]{0.54}& \makecell[c]{2.93}\\
    \makecell[c]{LPRE} & \makecell[c]{70.06} & \makecell[c]{\textbf{1.01}}\\
    \makecell[c]{SP} & \makecell[c]{\textbf{89.05}} & \makecell[c]{1.39} \\
    \makecell[c]{FMP-1} & \makecell[c]{\textbf{89.05}} & \makecell[c]{1.39}  \\
    \makecell[c]{FMP-2} & \makecell[c]{\textbf{89.07}} & \makecell[c]{1.39}  \\
    \hline
\end{tabular}
\end{center}
\label{tab:image-classification}
\vspace{-0.8cm}
\end{table}

\begin{table}[t]
\caption{{Testing Performance of Object Reorientation}}
\vspace{-0.3cm}
\label{table:1}
\begin{center}
\begin{tabular}{p{1.8cm}|p{1.5cm}|p{1.5cm}|p{1.5cm}}
    \hline
    \makecell[c]{Method} & \makecell[c]{tsr/\%} & \makecell[c]{100/pc} & \makecell[c]{tns} \\
    \hline
    \makecell[c]{RE} & \makecell[c]{77.50} & \makecell[c]{3.07} & \makecell[c]{2.18} \\
    \makecell[c]{LPRE} & \makecell[c]{79.33} & \makecell[c]{3.48}& \makecell[c]{1.90}\\
    \makecell[c]{SP} & \makecell[c]{81.64} & \makecell[c]{3.25} & \makecell[c]{2.05}\\
    \makecell[c]{FMP-1}& \makecell[c]{86.96} & \makecell[c]{3.91} & \makecell[c]{1.77}\\
    \makecell[c]{FMP-2}& \makecell[c]{{\bf 89.37}} & \makecell[c]{{\bf 4.55}} & \makecell[c]{{\bf 1.61}}\\
    \hline
\end{tabular}
\end{center}
\label{tab:reorientation}
\vspace{-0.4cm}
\end{table}

\subsection{Results}

\textbf{Comparisons to Baselines.} First, we compare our method with baselines in three tasks. For sequential image classification, we evaluate each method with the validation dataset of ImageNet~\cite{deng2009imagenet}. Note that in this task we regard the output of $\pi_{0}$ as the observation feature embedding (more details can be found in Appendix). That is, \textbf{SP} and \textbf{FMP-1} have the same settings for this task, thus reporting the same performances. We can see from Table \ref{tab:image-classification} that except for \textbf{RE}, \textbf{LPRE} shows the worst performance, while other three methods demonstrate similar performances. Referring to the analysis in \cite{ghosh2021generalization}, which proves that \textbf{SP} is an optimal policy for the sequential image classification task, our experimental results further figure out that \textbf{FMP-1} and \textbf{FMP-2} can also achieve optimal performance for this task.

For the object reorientation experiments, we present an additional metric ``pc'' in the reciprocal form, since the path planning cost will be infinite if the planning fails. Each method is evaluated with 207 unseen samples. Results in Table \ref{tab:reorientation} show that \textbf{FMP-2} outperforms other methods across all metrics, followed by \textbf{FMP-1}, which demonstrates that integrating the previous failure memory into policy training endows better policy tune-up during online testing and better generalization performance. Also, \textbf{FMP-2} reports better performance than \textbf{FMP-1} across all metrics. This might be due to the fact that, in this task, there exist some candidate reorientation poses which are similar to each other, thus leading to a similar point cloud feature. However, similar poses do not mean similar task assessment results. For example, flipping an object will fail due to the collision between the gripper and the table. But if the object is with a similar pose which leaves small space for the gripper, such manipulation might succeed. For \textbf{FMP-1}, it utilizes previous failures by conducting feature masking, which might hinder some possible successful actions. Moreover, randomly choosing actions among the remaining ones (\textbf{LPRE}), or applying memory as a mask (\textbf{SP}) neglects the dependency between the failure trials and the remaining actions, which shows lower performance compared to our policies. Besides, we can find that the performance \textbf{RE} is not too bad due to the small action set of this task.

\begin{table}[t]
\caption{{Testing Performance of Localization in Synthetic Environments.}}
\label{table:1}
\vspace{-0.3cm}
\begin{center}
\begin{tabular}{p{1.8cm}|p{1.5cm}|p{1.5cm}}
    \hline
    \makecell[c]{Method} & \makecell[c]{tsr/\%} & \makecell[c]{tns} \\
    \hline
    \makecell[c]{LPRE} & \makecell[c]{83.95} & \makecell[c]{\textbf{1.01}}\\
    \makecell[c]{SP} & \makecell[c]{84.23} & \makecell[c]{\textbf{1.01}} \\
    \makecell[c]{FMP-1} &\makecell[c]{\textbf{94.53}} & \makecell[c]{1.47}  \\
    \makecell[c]{FMP-2} & \makecell[c]{85.02} &  \makecell[c]{1.04}  \\
    \hline
\end{tabular}
\end{center}
\label{tab:localization-syn}
\vspace{-0.8cm}
\end{table}

\begin{figure}[t]
  \centering
  \includegraphics[width=\linewidth]{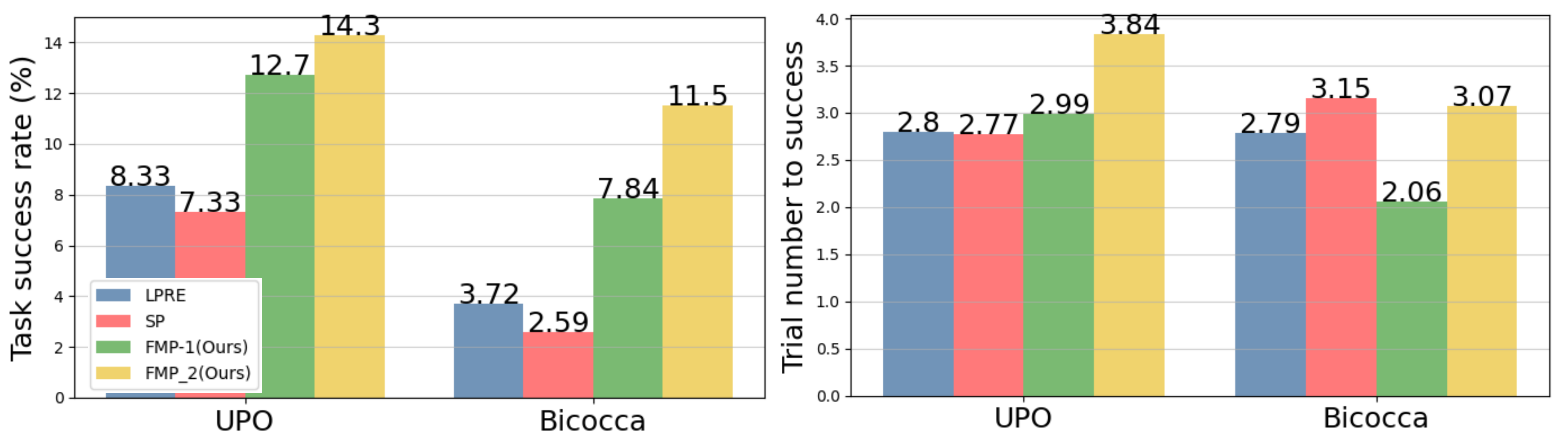}
  \vspace{-0.6cm}
  \caption{Testing Performance of Localization in Real Environments.} 
  \vspace{-0.6cm}
  \label{fig:localization-real}
\end{figure}

As for the localization task, the testing localization samples include three testing sequences in synthetic environments \cite{chen2021deep} and two testing sequences in real environments. Evaluation results in synthetic environments are shown in Table~\ref{tab:localization-syn} ($k=15$, more detailed results and ablation studies on $k$ can be found in Appendix). It is obvious that \textbf{FMP-1} achieves the best performance in the episode success rate with an average trial number less than 2. This large performance margin might come from the feature masking process, which hinders the similar feature of failed actions and encourages the policy to jump from the previous choices. \textbf{FMP-2} also shows better task success rate compared to \textbf{SP}. Note that the trial numbers to success of \textbf{LPRE} and \textbf{SP} are both close to 1, which indicates that these two policies cannot conduct effective adjustment by the self-assessment results. Fig.~\ref{fig:localization-real} demonstrates the testing results in real environments, which shows that our policies have better task success rate than \textbf{SP} and \textbf{LPRE} in all real environments. Overall, \textbf{FMP-2} shows better episode success rate, while \textbf{FMP-1} costs less trials to locate the position. This might be because the complex geometry of real-world maps calls for exploration in a small scope, which is the advantage of {\bf FMP-2}, while {\bf FMP-1} might hinder these similar positions if one of them fails. However, by jumping from the previous failure zone, {\bf FMP-1} is able to achieve success in less trials.
Note that we use $k=15$ because the complex geometry of real-world maps leads to multi-modal predictions, and we choose the action with maximum affordance in our experiments.

\begin{figure*}[t]
  \centering
  \includegraphics[width=0.83\linewidth]{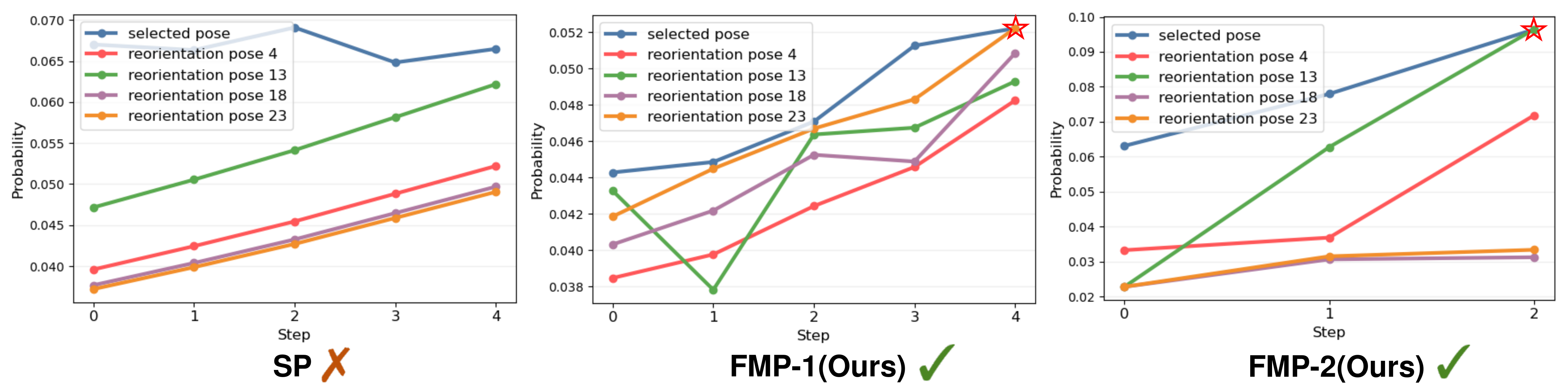}
  \vspace{-0.3cm}
  \caption{A Testing case in reorientation task of three policies. The setting of this case is: magic clean, initial pose: (-0.16, 0.16, 0.03, 1.59, 0.01, -3.14), target pose: (-0.15, 0.09, 0.13, 0.0, 0.0, 0.0), feasible reorientation poses indexes: (4, 13, 18, 23). We plot the normalized probability distributions of feasible poses the selected pose at all trial steps, and the stars label successful trials. \Checkmark means that the policy successfully finds the right pose, while \XSolidBrush means a failure. The decision sequences of three policies are \textbf{SP}: 12→7→2→17→8, \textbf{FMP-1}: 22→15→3→2→23, \textbf{FMP-2}: 24→17→13.} 
  \vspace{-0.6cm}
\label{fig:case-reorientation}
\end{figure*}

\textbf{Case Studies.} Fig.~\ref{fig:case-localization} presents some testing cases in the localization task of three policies. Since \textbf{SP} does not change its original distribution, all its decisions depend on the distribution generated from $\pi_{0}$. Instead, \textbf{FMP-1} concerns more on the feature correlation. When aware of a failed action, it can hinder the similar feature, thus jumping out of the previous failure zone. Also, by leveraging the recurrent implicit memory, \textbf{FMP-2} is also capable of adjusting the decisions, but shows a more conservative exploration process. Also, we can see the normalized probability changes of all feasible reorientation poses in an object reorientation case in Fig.~\ref{fig:case-reorientation}. In this task, the feasible poses are not unique, and the similar poses (with near pose indexes) do not mean similar feasibility. In this task, \textbf{FMP-2} performs better because it tends to explore the near pose first to confirm its feasibility. More case studies are shown in Appendix.
\subsection{Discussion}
Considering the architecture designs of our models and all the experimental results, we further analyze the advantages of our method, and how to choose an optimal policy in a specific tasks:
\begin{figure*}[t]
  \centering
  \includegraphics[width=0.83\linewidth]{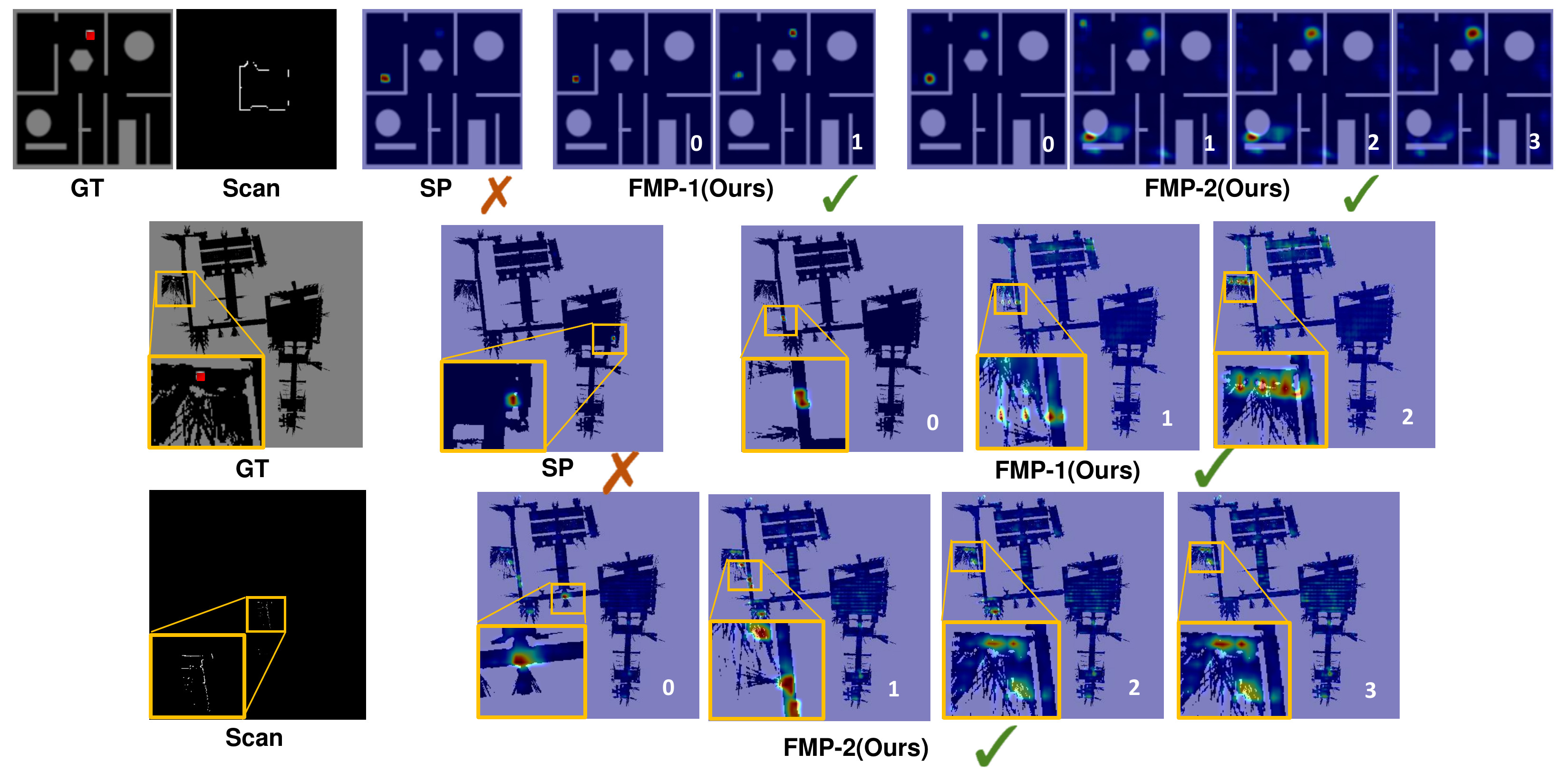}
  \vspace{-0.4cm}
  \caption{Testing cases in localization task of three policies.  The top row is a case in a synthetic environment. The left two columns show the global map with the ground truth position labeled as a red point, and the scan observation. Other columns show the prediction process and the distributions of three policies. The remaining row shows a case in the UPO dataset, where the keys are zoomed in with yellow boxes. The distribution is reflected by the color, where the value comes larger as the color comes closer to red. \Checkmark means that the policy successfully find the right position, while \XSolidBrush means a failure. } 
  \label{fig:case-localization}
  \vspace{-0.2cm}
\end{figure*}

{\bf What are the advantages of two policy architectures?} The same advantage of these two policy architectures is learnable. By integrating the self-assessment results into the policy training, \textbf{FMP-1} and \textbf{FMP-2} acquire the awareness of the previous failures, and have the capability of trimming the policy distribution according to these failures. However, these two architectures show different properties. \textbf{FMP-1} concerns more on the feature correlation, and tends to hinder the actions with similar features to that of the previously failed choices. Thus, \textbf{FMP-1} can achieve better performance in tasks where similar features ({\it e.g.} pose, geometry, visual attribute, and etc) lead to similar self-assessment results. Instead, \textbf{FMP-2} pays more attention to the recurrent memory, which encodes the observation feature and the previous trials. Consequently, \textbf{FMP-2} conducts a more conservative exploration process than \textbf{FMP-1}, and demonstrates better performances in tasks requiring an adjustment in a small scope.

{\bf What kind of policies is optimal?} For a task where similar features lead to similar self-assessment results, \textbf{FMP-1} can help jump out of local minimum, while for a task which needs to adjust in a small scope of the initial action, \textbf{FMP-2} shows better exploration strategy. Also, analyzing the results of the three tasks, we figure out that, when the actions are correlated, our method outperforms the process-of-elimination strategy. That is, {\it when the actions are correlated, the equality is broken between the action with the next-highest affordance and the action with the highest affordance conditioned on previous failures}. For sequential image classification, there is little correlation among different class choices. And the training data is adequate for fitting the distributions of all classes. Therefore, a simple process-of-elimination strategy can achieve optimal performance. Instead, as the action correlation increases, integrating action correlation into the policy learning endows better performance. Hence, we can achieve better performances in the localization task, and show the biggest advantage in the object reorientation task, whose actions have the highest correlation.

\subsection{Conclusion and Limitation}
\label{sec:conclusion}
In this paper, we propose to integrate the self-assessment results to learn a failure-aware policy, and propose two policy architectures. Experiments in three self-assessable robotics tasks demonstrate that our method outperforms other methods with higher task success rate with less trials. Moreover, we find that the action correlation has a large impact on the effect of our algorithm. The main limitation of our method lies in the assumption of the discrete action set. This limitation comes from the construction of the representation $m_t$. In our paper, it is represented as a finite matrix, of which each element corresponds an action. In future work, more general representations of $m_t$ for continuous actions can be studied. Also, in this paper, we assume that the self-assessment module can accurately predict failure, and $\pi_0$ is a learned differentiable policy with a feature bottleneck layer. Further works can involve the assessment uncertainty in more real-world applications, and extend to more general $\pi_0$. 




\clearpage
\bibliographystyle{IEEEtran}
\bibliography{IEEEabrv,ref}

\begin{thebibliography}{10}
\providecommand{\url}[1]{#1}
\csname url@samestyle\endcsname
\providecommand{\newblock}{\relax}
\providecommand{\bibinfo}[2]{#2}
\providecommand{\BIBentrySTDinterwordspacing}{\spaceskip=0pt\relax}
\providecommand{\BIBentryALTinterwordstretchfactor}{4}
\providecommand{\BIBentryALTinterwordspacing}{\spaceskip=\fontdimen2\font plus
\BIBentryALTinterwordstretchfactor\fontdimen3\font minus
  \fontdimen4\font\relax}
\providecommand{\BIBforeignlanguage}[2]{{%
\expandafter\ifx\csname l@#1\endcsname\relax
\typeout{** WARNING: IEEEtran.bst: No hyphenation pattern has been}%
\typeout{** loaded for the language `#1'. Using the pattern for}%
\typeout{** the default language instead.}%
\else
\language=\csname l@#1\endcsname
\fi
#2}}
\providecommand{\BIBdecl}{\relax}
\BIBdecl

\bibitem{ghosh2021generalization}
D.~Ghosh, J.~Rahme, A.~Kumar, A.~Zhang, R.~P. Adams, and S.~Levine, ``Why
  generalization in rl is difficult: Epistemic pomdps and implicit partial
  observability,'' \emph{Advances in Neural Information Processing Systems},
  vol.~34, 2021.

\bibitem{isele2018safe}
D.~Isele, A.~Nakhaei, and K.~Fujimura, ``Safe reinforcement learning on
  autonomous vehicles,'' in \emph{2018 IEEE/RSJ International Conference on
  Intelligent Robots and Systems (IROS)}.\hskip 1em plus 0.5em minus
  0.4em\relax IEEE, 2018, pp. 1--6.

\bibitem{srinivasan2020learning}
K.~Srinivasan, B.~Eysenbach, S.~Ha, J.~Tan, and C.~Finn, ``Learning to be safe:
  Deep rl with a safety critic,'' \emph{arXiv preprint arXiv:2010.14603}, 2020.

\bibitem{krasowski2020safe}
H.~Krasowski, X.~Wang, and M.~Althoff, ``Safe reinforcement learning for
  autonomous lane changing using set-based prediction,'' in \emph{2020 IEEE
  23rd International Conference on Intelligent Transportation Systems
  (ITSC)}.\hskip 1em plus 0.5em minus 0.4em\relax IEEE, 2020, pp. 1--7.

\bibitem{mokhtari2021safe}
K.~Mokhtari and A.~R. Wagner, ``Safe deep q-network for autonomous vehicles at
  unsignalized intersection,'' \emph{arXiv preprint arXiv:2106.04561}, 2021.

\bibitem{lin2018efficient}
K.~Lin, R.~Zhao, Z.~Xu, and J.~Zhou, ``Efficient large-scale fleet management
  via multi-agent deep reinforcement learning,'' in \emph{Proceedings of the
  24th ACM SIGKDD International Conference on Knowledge Discovery \& Data
  Mining}, 2018, pp. 1774--1783.

\bibitem{chen2021multi}
D.~Chen, Z.~Li, Y.~Wang, L.~Jiang, and Y.~Wang, ``Deep multi-agent
  reinforcement learning for highway on-ramp merging in mixed traffic,''
  \emph{arXiv preprint arXiv:2105.05701}, 2021.

\bibitem{hochreiter1997long}
S.~Hochreiter and J.~Schmidhuber, ``Long short-term memory,'' \emph{Neural
  computation}, vol.~9, no.~8, pp. 1735--1780, 1997.

\bibitem{cho2014learning}
K.~Cho, B.~Merrienboer, C.~Gulcehre, F.~Bougares, H.~Schwenk, and Y.~Bengio,
  ``Learning phrase representations using rnn encoder-decoder for statistical
  machine translation,'' in \emph{EMNLP}, 2014.

\bibitem{burghouts2020robotic}
G.~J. Burghouts, A.~Huizing, and M.~A. Neerincx, ``Robotic self-assessment of
  competence,'' \emph{ACM/IEEE International Conference on Human-Robot
  Interaction (HRI)}, 2020.

\bibitem{nortonmetrics}
A.~Norton, H.~Admoni, J.~Crandall, T.~Fitzgerald, A.~Gautam, M.~Goodrich,
  A.~Saretsky, M.~Scheutz, R.~Simmons, A.~Steinfeld \emph{et~al.}, ``Metrics
  for robot proficiency self-assessment and communication of proficiency in
  human-robot teams,'' \emph{ACM Transactions on Human-Robot Interaction},
  2022.

\bibitem{deptula2019approximate}
P.~Deptula, H.-Y. Chen, R.~A. Licitra, J.~A. Rosenfeld, and W.~E. Dixon,
  ``Approximate optimal motion planning to avoid unknown moving avoidance
  regions,'' \emph{IEEE Transactions on Robotics}, vol.~36, no.~2, pp.
  414--430, 2019.

\bibitem{xu2021efficient}
K.~Xu, H.~Yu, R.~Huang, D.~Guo, Y.~Wang, and R.~Xiong, ``Efficient object
  manipulation to an arbitrary goal pose: Learning-based anytime prioritized
  planning,'' \emph{2022 IEEE International Conference on Robotics and
  Automation (ICRA)}, 2022.

\bibitem{kumar2021error}
V.~Kumar, S.~Ha, and C.~K. Liu, ``Error-aware policy learning: Zero-shot
  generalization in partially observable dynamic environments,''
  \emph{Robotics: Science and Systems (RSS)}, 2021.

\bibitem{finn2017deep}
C.~Finn and S.~Levine, ``Deep visual foresight for planning robot motion,'' in
  \emph{2017 IEEE International Conference on Robotics and Automation
  (ICRA)}.\hskip 1em plus 0.5em minus 0.4em\relax IEEE, 2017, pp. 2786--2793.

\bibitem{ebert2018visual}
F.~Ebert, C.~Finn, S.~Dasari, A.~Xie, A.~Lee, and S.~Levine, ``Visual
  foresight: Model-based deep reinforcement learning for vision-based robotic
  control,'' \emph{arXiv preprint arXiv:1812.00568}, 2018.

\bibitem{wang2019learning}
A.~Wang, T.~Kurutach, K.~Liu, P.~Abbeel, and A.~Tamar, ``Learning robotic
  manipulation through visual planning and acting,'' in \emph{Robotics: science
  and systems}, 2019.

\bibitem{di2020safari}
N.~Di~Palo and E.~Johns, ``Safari: Safe and active robot imitation learning
  with imagination,'' \emph{arXiv preprint arXiv:2011.09586}, 2020.

\bibitem{xu2021push}
K.~Xu, H.~Yu, Q.~Lai, Y.~Wang, and R.~Xiong, ``Efficient learning of
  goal-oriented push-grasping synergy in clutter,'' \emph{IEEE Robotics and
  Automation Letters}, vol.~6, no.~4, pp. 6337--6344, 2021.

\bibitem{xu2021learning}
Z.~Xu, Z.~He, J.~Wu, and S.~Song, ``Learning 3d dynamic scene representations
  for robot manipulation,'' in \emph{Conference on Robot Learning}.\hskip 1em
  plus 0.5em minus 0.4em\relax PMLR, 2021, pp. 126--142.

\bibitem{richter2017safe}
C.~Richter and N.~Roy, ``Safe visual navigation via deep learning and novelty
  detection,'' 2017.

\bibitem{wellhausen2020safe}
L.~Wellhausen, R.~Ranftl, and M.~Hutter, ``Safe robot navigation via
  multi-modal anomaly detection,'' \emph{IEEE Robotics and Automation Letters},
  vol.~5, no.~2, pp. 1326--1333, 2020.

\bibitem{kahn2017uncertainty}
G.~Kahn, A.~Villaflor, V.~Pong, P.~Abbeel, and S.~Levine, ``Uncertainty-aware
  reinforcement learning for collision avoidance,'' \emph{arXiv preprint
  arXiv:1702.01182}, 2017.

\bibitem{lutjens2019safe}
B.~L{\"u}tjens, M.~Everett, and J.~P. How, ``Safe reinforcement learning with
  model uncertainty estimates,'' in \emph{2019 International Conference on
  Robotics and Automation (ICRA)}.\hskip 1em plus 0.5em minus 0.4em\relax IEEE,
  2019, pp. 8662--8668.

\bibitem{wong2022error}
J.~Wong, A.~Tung, A.~Kurenkov, A.~Mandlekar, L.~Fei-Fei, S.~Savarese, and
  R.~Mart{\'\i}n-Mart{\'\i}n, ``Error-aware imitation learning from
  teleoperation data for mobile manipulation,'' in \emph{Conference on Robot
  Learning}.\hskip 1em plus 0.5em minus 0.4em\relax PMLR, 2022, pp. 1367--1378.

\bibitem{martinez2007active}
R.~Martinez-Cantin, N.~de~Freitas, A.~Doucet, and J.~A. Castellanos, ``Active
  policy learning for robot planning and exploration under uncertainty.'' in
  \emph{Robotics: Science and systems}, vol.~3, 2007, pp. 321--328.

\bibitem{peters2010relative}
J.~Peters, K.~Mulling, and Y.~Altun, ``Relative entropy policy search,'' in
  \emph{Twenty-Fourth AAAI Conference on Artificial Intelligence}, 2010.

\bibitem{sutton2018reinforcement}
R.~S. Sutton and A.~G. Barto, \emph{Reinforcement learning: An
  introduction}.\hskip 1em plus 0.5em minus 0.4em\relax MIT press, 2018.

\bibitem{houthooft2016vime}
R.~Houthooft, X.~Chen, Y.~Duan, J.~Schulman, F.~De~Turck, and P.~Abbeel,
  ``Vime: Variational information maximizing exploration,'' \emph{Advances in
  neural information processing systems}, vol.~29, 2016.

\bibitem{bellemare2016unifying}
M.~Bellemare, S.~Srinivasan, G.~Ostrovski, T.~Schaul, D.~Saxton, and R.~Munos,
  ``Unifying count-based exploration and intrinsic motivation,'' \emph{Advances
  in neural information processing systems}, vol.~29, 2016.

\bibitem{pathak2017curiosity}
D.~Pathak, P.~Agrawal, A.~A. Efros, and T.~Darrell, ``Curiosity-driven
  exploration by self-supervised prediction,'' in \emph{International
  conference on machine learning}.\hskip 1em plus 0.5em minus 0.4em\relax PMLR,
  2017, pp. 2778--2787.

\bibitem{tang2017exploration}
H.~Tang, R.~Houthooft, D.~Foote, A.~Stooke, O.~Xi~Chen, Y.~Duan, J.~Schulman,
  F.~DeTurck, and P.~Abbeel, ``\# exploration: A study of count-based
  exploration for deep reinforcement learning,'' \emph{Advances in neural
  information processing systems}, vol.~30, 2017.

\bibitem{burda2018exploration}
Y.~Burda, H.~Edwards, A.~Storkey, and O.~Klimov, ``Exploration by random
  network distillation,'' \emph{arXiv preprint arXiv:1810.12894}, 2018.

\bibitem{machado2020count}
M.~C. Machado, M.~G. Bellemare, and M.~Bowling, ``Count-based exploration with
  the successor representation,'' in \emph{Proceedings of the AAAI Conference
  on Artificial Intelligence}, vol.~34, no.~04, 2020, pp. 5125--5133.

\bibitem{whitney2021decoupled}
W.~F. Whitney, M.~Bloesch, J.~T. Springenberg, A.~Abdolmaleki, K.~Cho, and
  M.~Riedmiller, ``Decoupled exploration and exploitation policies for
  sample-efficient reinforcement learning,'' \emph{arXiv preprint
  arXiv:2101.09458}, 2021.

\bibitem{deng2009imagenet}
J.~Deng, W.~Dong, R.~Socher, L.-J. Li, K.~Li, and L.~Fei-Fei, ``Imagenet: A
  large-scale hierarchical image database,'' in \emph{2009 IEEE conference on
  computer vision and pattern recognition}.\hskip 1em plus 0.5em minus
  0.4em\relax Ieee, 2009, pp. 248--255.

\bibitem{xiang2020sapien}
F.~Xiang, Y.~Qin, K.~Mo, Y.~Xia, H.~Zhu, F.~Liu, M.~Liu, H.~Jiang, Y.~Yuan,
  H.~Wang \emph{et~al.}, ``Sapien: A simulated part-based interactive
  environment,'' in \emph{Proceedings of the IEEE/CVF Conference on Computer
  Vision and Pattern Recognition}, 2020, pp. 11\,097--11\,107.

\bibitem{chen2021deep}
R.~Chen, H.~Yin, Y.~Jiao, G.~Dissanayake, Y.~Wang, and R.~Xiong, ``Deep
  samplable observation model for global localization and kidnapping,''
  \emph{IEEE Robotics and Automation Letters}, vol.~6, no.~2, pp. 2296--2303,
  2021.

\bibitem{ramon2014navigating}
R.~Ram{\'o}n-Vigo, J.~P{\'e}rez, F.~Caballero, and L.~Merino, ``Navigating
  among people in crowded environment: Datasets for localization and human
  robot interaction,'' in \emph{Proceedings of the Workshop on Robots in
  Clutter: Perception and Interaction in Clutter, IEEE/RSJ International
  Conference on Intelligent Robots and Systems (IROS)}.\hskip 1em plus 0.5em
  minus 0.4em\relax Citeseer, 2014.

\bibitem{bonarini2006rawseeds}
A.~Bonarini, W.~Burgard, G.~A.~E. Fontana, M.~Matteucci, D.~Sorrenti, and
  J.~Tardos, ``Rawseeds: Robotics advancement through web-publishing of
  sensorial and elaborated extensive data sets,'' in \emph{Workshop on
  Benchmarks in Robotics Research at IEEE/RSJ International Conference on
  Intelligent Robots and Systems (IROS 2006)}, 2006, pp. 1--5.

\bibitem{ceriani2009rawseeds}
S.~Ceriani, G.~Fontana, A.~Giusti, D.~Marzorati, M.~Matteucci, D.~Migliore,
  D.~Rizzi, D.~G. Sorrenti, and P.~Taddei, ``Rawseeds ground truth collection
  systems for indoor self-localization and mapping,'' \emph{Autonomous Robots},
  vol.~27, no.~4, pp. 353--371, 2009.

\bibitem{he2016deep}
K.~He, X.~Zhang, S.~Ren, and J.~Sun, ``Deep residual learning for image
  recognition,'' in \emph{Proceedings of the IEEE conference on computer vision
  and pattern recognition}, 2016, pp. 770--778.

\bibitem{rw2019timm}
R.~Wightman, ``Pytorch image models,''
  \url{https://github.com/rwightman/pytorch-image-models}, 2019.

\bibitem{mnih2015human}
V.~Mnih, K.~Kavukcuoglu, D.~Silver, A.~A. Rusu, J.~Veness, M.~G. Bellemare,
  A.~Graves, M.~Riedmiller, A.~K. Fidjeland, G.~Ostrovski \emph{et~al.},
  ``Human-level control through deep reinforcement learning,'' \emph{nature},
  vol. 518, no. 7540, pp. 529--533, 2015.

\end{thebibliography}

\clearpage
\appendices

\section{Experiment Implementation Details}
In this section, we will introduce three typical robotics tasks (shown in Fig.~\ref{fig:task}) that we use in experiment evaluation, and demonstrate how to apply our algorithm in these specific tasks.

\subsection{Sequential Image Classification} 
\label{sec:exp-setup-1}
\textbf{Task Definition.} Our first task is sequential image classification, which is motivated by \cite{ghosh2021generalization} and the setting is also similar. In this task, the robot observes an image from the dataset at the beginning of an episode and identifies a label for this image. After choosing a label, the robot will receive feedback from the self-assessment module about whether the choice is correct. In real applications, the assessment process can be conducted by human-robot interaction. If incorrect, the robot is supposed to re-choose a label until evaluated as correct by the self-assessment module.

\textbf{Policy Architecture Implementation.} In this task, we use the pre-trained model of ResNet18~\cite{he2016deep} on ImageNet~\cite{deng2009imagenet} from \cite{rw2019timm}
as the learned policy $\pi_{0}$. Self-assessment representation $m_t$ is a normalized vector with the same size as the output of $\pi_{0}$, which is initialized as the softmax output of $\pi_{0}$. And for \textbf{FMP-1}, we use $\pi_{0}$ ({\it i.e.} ResNet18) as the observation encoder, whose output can be regarded as the observation embedding feature. And the memory encoder and the decoder are identity transforms. Thus, such an implementation of \textbf{FMP-1} is equivalent to that of process-of-elimination \cite{ghosh2021generalization} ({\it i.e.} first choosing the action with the highest affordance, if incorrect, then the second, and so forth). For \textbf{FMP-2}, we use $\pi_{0}$ ({\it i.e.} ResNet18) as the observation encoder and a GRU~\cite{cho2014learning} as the memory-aware module. Memory encoder and decoder are both identity transforms. 

\textbf{Training and Testing Details.} We apply DQN \cite{mnih2015human} to train $\pi_{\mathrm{FA}}$ with shared parameters of $\pi_{0}$ fixed. For each episode, the robot is provided with an image from the ImageNet training set, and gives a sequence of guesses for the image label among the total 1000 labels. At training time, if the selected label is correct, the robot gets a reward of $r = 1$, and the episode ends. Otherwise, the robot gets a reward of $r = 0$, and the episode continues to the next time step. We limit the trial times up to $t=5$. We train the network with SGD optimizer with L1 loss, using learning rate of $10^{-4}$, momentum of 0.9 and weight decay $2^{-5}$, and the future discount $\gamma$ is constant at 0.2. At testing time, we evaluate the trained policy with the validation set of ImageNet~(50k images) with the same episode setting as training.

\subsection{Object Reorientation}
\textbf{Task Definition.} Object reorientation is a manipulation task where a robot is supposed to choose a reorientation object pose to achieve a feasible or even optimal pick-reorient-place process \cite{xu2021efficient}. In this task, the robot is provided with an assigned object, with its mesh model, initial pose and target pose, and is supposed to choose a reorientation pose as a transition since the one-step pick-and-place might be failed due to collision between the robot and the environment. The policy in \cite{xu2021efficient} first generates a finite set of reorientation pose candidates and predicts an affordance map of these poses, and the one with the highest affordance will be conducted. Also, pre-acting in a simulator with the planning algorithm RRT predicts the path planning cost of the whole pick-reorient-place process after choosing a reorientation pose, which serves as the self-assessment module. In this task, we set the reorientation pose candidate number $n=25$.

\begin{figure*}[t]
  \centering
  \includegraphics[width=\linewidth]{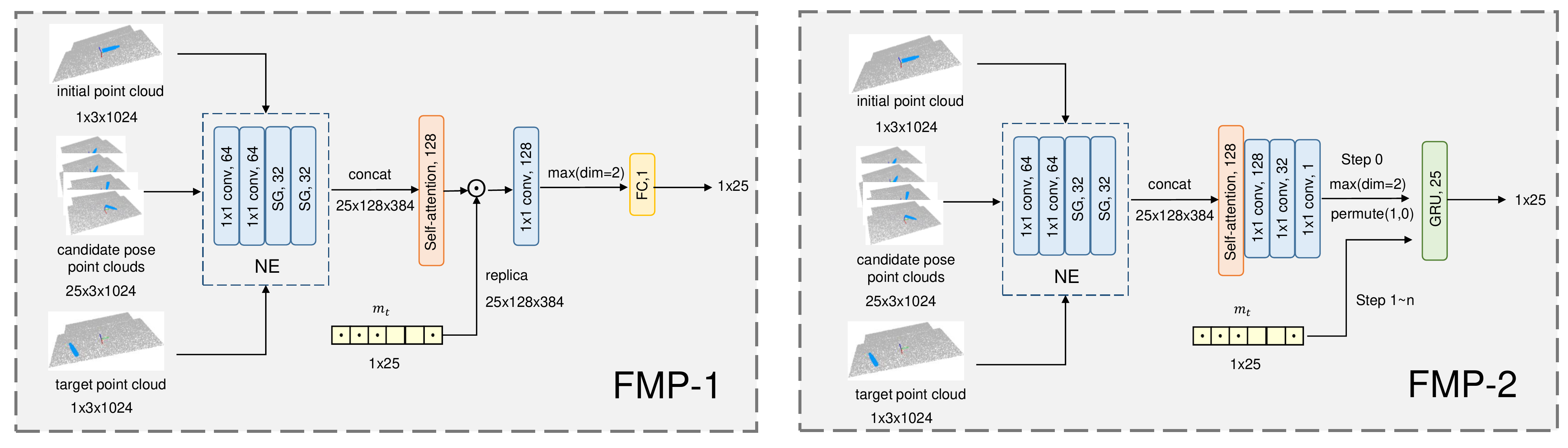}
  \caption{Network architectures of reorientation task. Note that SG is sampling and grouping layers introduced in \href{PCT}{https://github.com/MenghaoGuo/PCT}.} 
  \label{fig:reorientation-network}
\end{figure*}

\textbf{Policy Architecture Implementation.} We use the policy in \cite{xu2021efficient} as the learned policy $\pi_{0}$. Self-assessment representation $m_t$ is a binary vector with the same size as the candidate pose set. For \textbf{FMP-1}, the feature extraction module (NE) of $\pi_{0}$ followed by a self-attention layer is used as the observation encoder, which outputs $n$ embedding concated features corresponding to $n$ pose candidates. The decoder is the evaluation module (PCEN) of $\pi_{0}$. And the memory encoder is a replica transform, which converts the shape of $m_t$ to the same as the observation feature. For \textbf{FMP-2}, the observation encoder is that of \textbf{FMP-1} followed by additional convolution blocks, the decoder and the memory encoder are identity transforms, and the memory-aware module is a GRU~\cite{cho2014learning}. The network architectures are presented in Fig.~\ref{fig:reorientation-network}.

\begin{figure*}[t]
  \centering
  \includegraphics[width=\linewidth]{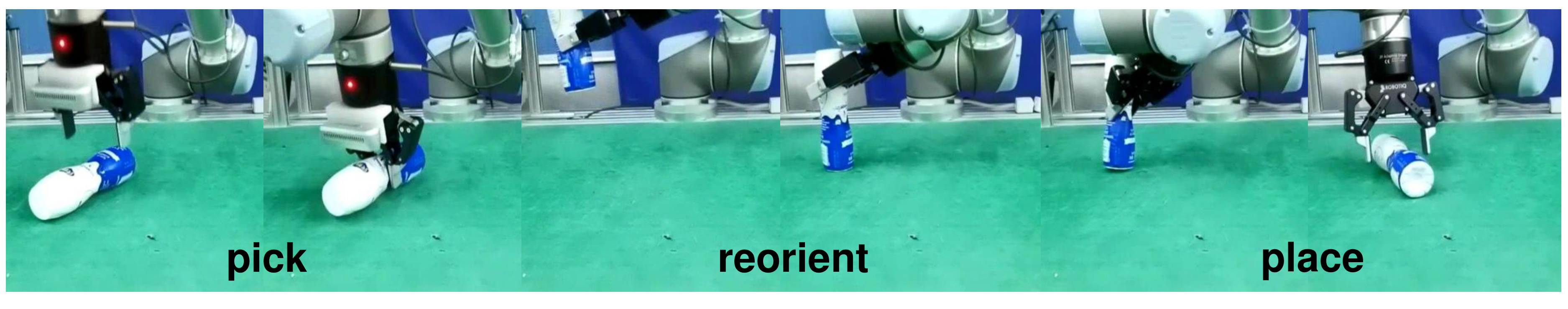}
  \caption{An example pick-reorient-place sequence in real-world environment.}
  \label{fig:reorientation-sequence}
\end{figure*}

\textbf{Training and Testing Details.} $\pi_{0}$ is pre-trained via behavior cloning with a dataset containing 3048 samples labeled with the path planning costs. To build this dataset, we collect each sample in SAPIEN~\cite{xiang2020sapien} with a UR5 arm by randomly choosing an object model from the model set consisting of 21 3D object models, with an initial pose and a target pose randomly sampled from the stable place poses. Then the robot traversely executes the planned pick-reorient-place trajectories of all the reorientation pose candidates, which generates the planning costs of all the ``actions''. The planning algorithm is RRT, and the planning cost is numerically dependent on the trajectory's existence and length. To train $\pi_{\mathrm{FA}}$, DQN \cite{mnih2015human} is applied with parameters of NE fixed. For each episode, a model is randomly sampled from the model set, with an initial pose and a goal pose randomly sampled from the stable place poses. Then the policy chooses a reorientation object pose and executes the planned trajectories. If planning fails, another pose will be selected until the planning succeeds or up to $t=5$ trials. The reward design is the same as that in \cite{xu2021efficient}. We train the network with Adam optimizer with Huber loss, using learning rate $10^{-4}$, weight decay $2^{-5}$, betas $(0.9. 0.99)$, and the future discount $\gamma$ is constant at 0.2. At testing time, we evaluate the trained policy with 207 unseen samples with the same episode setting. The real-world environment contains a UR5 arm, and an example sequence is shown in Fig.~\ref{fig:reorientation-sequence}.
\subsection{Localization}
\begin{figure}[t]
  \centering
  \subfigure[UPO]{
    \label{fig:map-upo} 
    \includegraphics[scale=0.10]{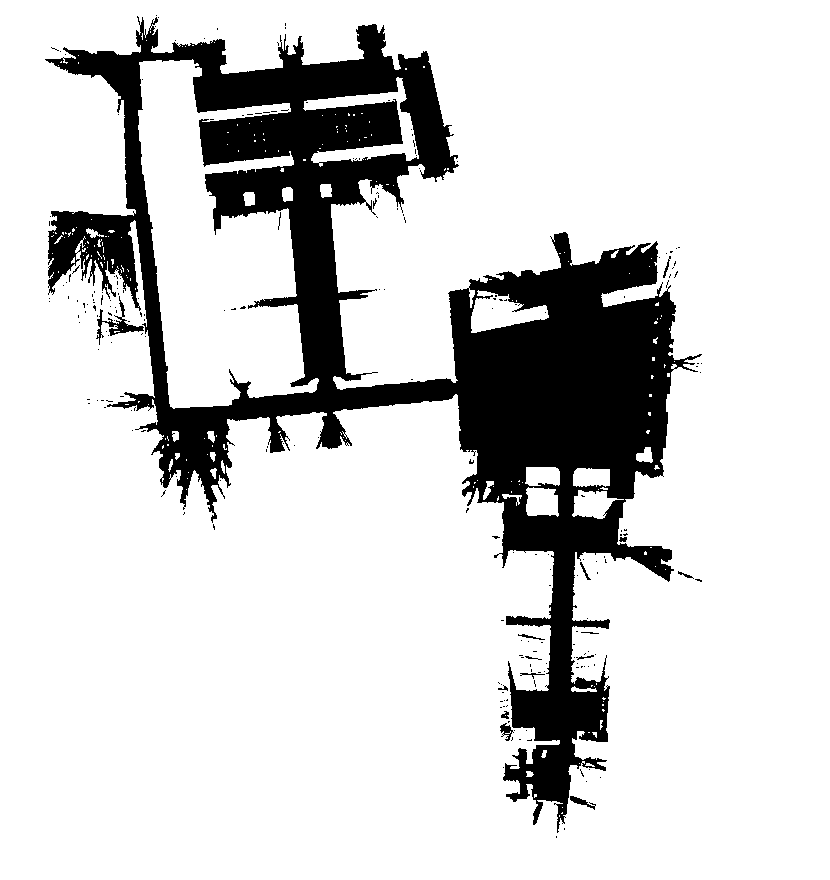}}
  \hspace{0.1in} 
  \subfigure[Bicocca]{
    \label{fig:map-bicocca1} 
    \includegraphics[scale=0.30]{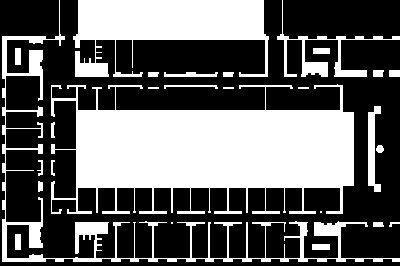}}
  \caption{Real-world global maps for the localization task.}
  \label{fig:maps} 
\end{figure}

\textbf{Task Definition.} We also evaluate our method with a typical mobile robotics task: localization. The task setting is the same as in \cite{chen2021deep}. In this task, the robot predicts its position given a global map and an observed scan. And the self-assessment module identifies the localization accuracy. In real applications, we can use some registration algorithms as self-assessment module which measures the localization accuracy. In our experiments, though, we use datasets to provide ground truth for convenience. In this task, if the predicted position is at the $k\times k$ neighborhood of the ground truth position, the action is regarded as successful. Otherwise, the robot will re-choose the position to reach the localization accuracy. Synthetic datasets and real-world datasets (UPO \cite{ramon2014navigating} and Bicocca \cite{bonarini2006rawseeds}\cite{ceriani2009rawseeds} shown in Fig.~\ref{fig:maps}) are used to train and test as the same way in \cite{chen2021deep}. Following \cite{chen2021deep}, Bicocca is split into two datasets. The original images are resized into $H\times W$ ($H=W=128$) before feeding into the network. 

\begin{figure*}[t]
  \centering
  \includegraphics[width=\linewidth]{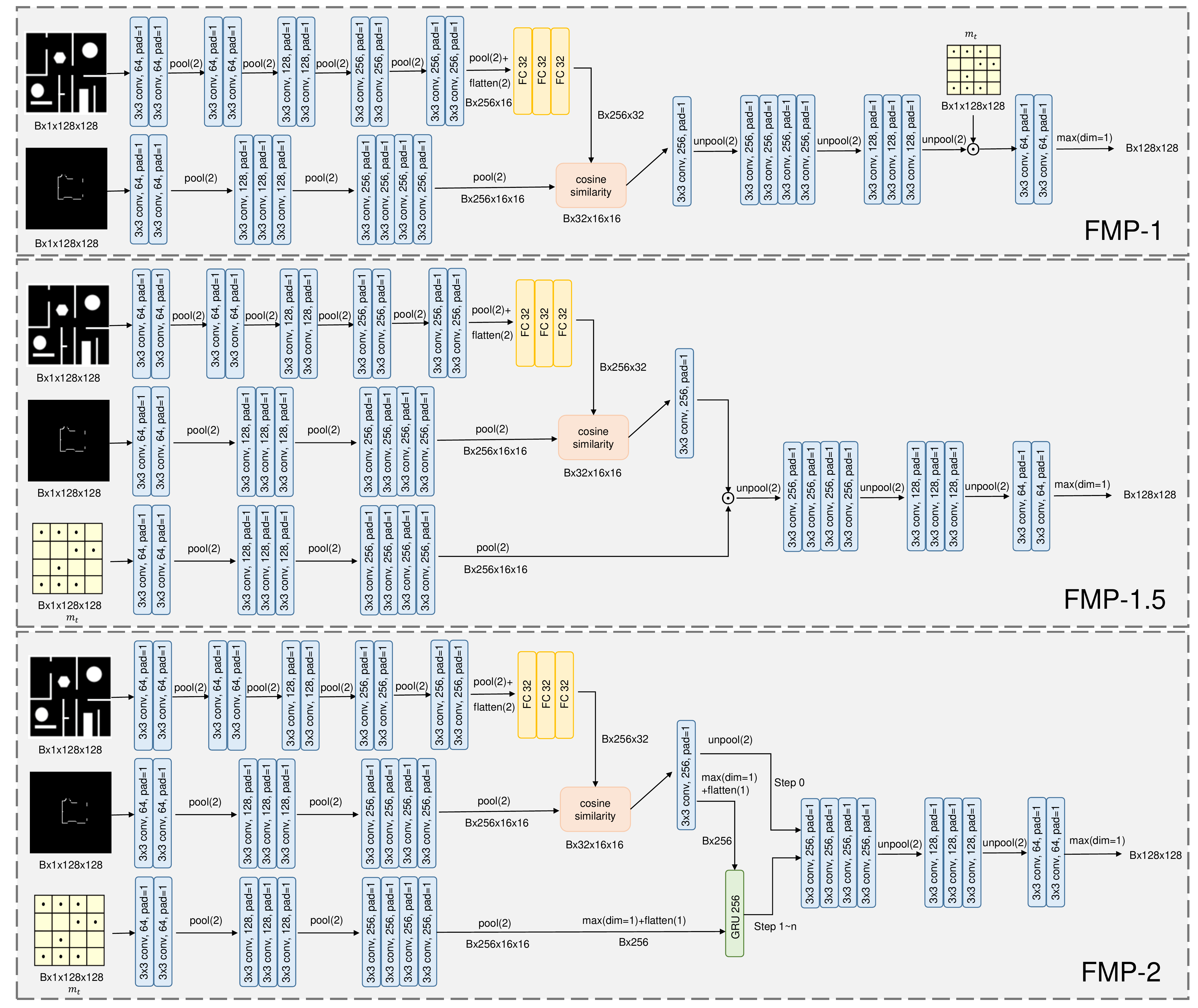}
  \caption{Network architectures of localization task.} 
  \label{fig:localization-network}
\end{figure*}

\textbf{Policy Architecture Implementation.} We use the pre-trained models in \cite{chen2021deep} as the learned policy $\pi_{0}$. Self-assessment representation $m_t$ is a binary mask with shape $H\times W$, which is the shape of the global map. $m_t$ is updated with neighborhood $k\times k$. For \textbf{FMP-1}, we use the encoder and part of the decoder of $\pi_{0}$ as the observation encoder and the remaining decoder of $\pi_{0}$ as the decoder. And memory encoder is an identity transform. For \textbf{FMP-2}, the observation encoder and the decoder are the same as $\pi_{0}$, with a separate memory encoder with the same network architecture as the observation encoder, and a GRU \cite{cho2014learning} as the memory-aware module. 
The network architectures are presented in Fig.~\ref{fig:localization-network}.

\textbf{Training and Testing Details.} We apply DQN \cite{mnih2015human} to train $\pi_{\mathrm{FA}}$ with parameters of observation encoder fixed. For each episode, a global map and an observed scan are fed into the policy, and an initial affordance map with the same size as the global map is predicted at the first step. If failed, $m_t$ will be updated and re-decision will be conducted by $\pi_{\mathrm{FA}}$ in the following steps. Successfully reaching the localization accuracy or coming up to the limited trial times ($t=5$) ends the episode. At training time, if the selected position is evaluated as successful, the robot gets a reward of $r = 1$, and the episode ends. Otherwise, the robot gets a reward of $r = 0$, and the episode continues to the next step. We train the network with Adam optimizer with smooth L1 loss, using learning rate of $10^{-3}$, weight decay $2^{-6}$, and the future discount $\gamma$ is constant at 0.2. At testing time, we evaluate the trained policy with the same episode setting as training. Three sequences of synthetic data in two unseen synthetic maps, and three sequences of real-world data in three real-world maps with unseen observation are used as the validation set.

\section{More Experimental Results}
\subsection{Ablation Studies}
\begin{table}[t]
\centering
\caption{Testing Performance of Localization in Synthetic Environments.}
\begin{threeparttable}
    \resizebox{\linewidth}{!}{
        \begin{tabular}{c|c|c|c|c|c|c|c|c|c}
        \toprule
        \multirow{2}{*}{Method} & \multirow{2}{*}{{\it k}} & \multicolumn{4}{c|}{tsr/\%} & \multicolumn{4}{c}{tns/\%} \\
        \cline{3-10}
         &  & S-1 & S-2 & S-3 & avg & S-1 & S-2 & S-3 & avg \\
        \hline
        LPRE & \multirow{4}{*}{15} & 91.54& 74.16& 86.15& 83.95& 1.03& {\bf 1.00}& {\bf 1.00} & {\bf 1.01}\\
        SP & & 90.77 & 75.00 & 86.92 & 84.23& {\bf 1.00}& 1.01 & 1.01 & {\bf 1.01}\\
        FMP-1 & & {\bf 98.46}& {\bf 86.67}& {\bf 98.46}& {\bf 94.53}& 1.22& 1.75 & 1.45 & 1.47\\
        FMP-2 & & 90.77& 75.83& 88.46& 85.02& 1.01& 1.09 & 1.03 & 1.04 \\
        \hline
        LPRE & \multirow{4}{*}{9} & 89.23& 74.16& 83.85& 82.41& {\bf 1.00}& {\bf 1.00}& {\bf 1.00} & {\bf 1.00}\\
        SP & & 90.77 & 75.00 & 85.38 & 83.72& 1.03& 1.01 & 1.05 & 1.03\\
        FMP-1 & & {\bf 91.54}& {\bf 84.59}& {\bf 93.85}& {\bf 89.99}& 1.08& 1.65 & 1.13 & 1.29\\
        FMP-2 & & 87.69& 79.17& 83.85&83.57 & 1.01& 1.01 & 1.02 &1.01\\
        \hline
        LPRE & \multirow{4}{*}{5} & 71.54& 52.50& 72.31& 65.45& {\bf 1.00}& {\bf 1.00}& {\bf 1.00} & {\bf 1.00}\\
        SP & & 84.61 & 69.17 & 81.54 & 78.44& 1.31& 1.37 & 1.14 & 1.27\\
        FMP-1 & & {\bf 86.15}& {\bf 76.67}& {\bf 91.54}& {\bf 84.79}& 1.21& 1.46 & 1.44 & 1.37\\
        FMP-2 & & 66.15& 59.17& 66.15& 63.82& 2.38& 1.17 & 1.01 & 1.08 \\
        \bottomrule
        \end{tabular}
        }
\vspace{0.5em}
\begin{tablenotes}
 \item[2]* S-1, S-2, S-3 represent three synthetic testing sequences.
\end{tablenotes}
\end{threeparttable}
\label{tab:localization-synthetic-detail}
\end{table}

\textbf{Localization Neighborhood Size.} We conduct an ablation study on the neighborhood size in the localization task (shown in Table.~\ref{tab:localization-syn}, where $k=5, 9, 15$). We can see that \textbf{FMP-1} shows the best performance across all neighborhood sizes. And the advantage becomes greater as localization accuracy increases ({\it i.e.} $k$ decreases). Instead, \textbf{FMP-2} shows better or comparable performances compared to \textbf{SP} when $k=9, 15$, but presents worst performance when $k=5$. This might be because \textbf{FMP-1} concerns more with the recurrent memory across the sequential decision process, thus tending to choose the position near the previously chosen ones. Under high localization accuracy requirements, such a strategy might be stuck in the local minimum. 
\begin{table}[t]
\centering
\caption{Testing Performance of Localization in Synthetic Environments with different implementations of policy architecture 1.}
\begin{threeparttable}
    \resizebox{\linewidth}{!}{
        \begin{tabular}{c|c|c|c|c|c|c|c|c}
        \toprule
        \multirow{2}{*}{Method} & \multicolumn{4}{c|}{tsr/\%} & \multicolumn{4}{c}{tns/\%} \\
        \cline{2-9}
         &  S-1 & S-2 & S-3 & avg & S-1 & S-2 & S-3& avg \\
        \hline
        FMP-1 & {\bf 98.46}& {86.67}& {\bf 98.46}& 94.53& 1.22& 1.75 & 1.45 &1.47\\
        FMP-1.5 & 96.92& {\bf 95.83}& 94.62& {\bf 95.79}&{\bf 1.13}& {\bf 1.60} & {\bf 1.21} &{\bf 1.31} \\
        \bottomrule
        \end{tabular}
        }
\vspace{0.5em}
\begin{tablenotes}
 \item[2]* S-1, S-2, S-3 represent three synthetic testing sequences.
\end{tablenotes}
\end{threeparttable}
\label{tab:localization-synthetic-fmp1.5}
\end{table}

\begin{table}[t]
\centering
\caption{Testing Performance of Localization in Real-world Environments with different implementations of policy architecture 1.}
\begin{threeparttable}
    \resizebox{0.6\linewidth}{!}{
        \begin{tabular}{c|c|c|c|c}
        \toprule
        \multirow{2}{*}{Method} & \multicolumn{2}{c|}{tsr/\%} & \multicolumn{2}{c}{tns/\%} \\
        \cline{2-5}
         &  U & B & U & B\\
        \hline
        FMP-1 & 12.67& {\bf 7.84}& {\bf 2.99}& {\bf 2.06} \\
        FMP-1.5 & {\bf 31.00}& 6.89& {\bf 2.99}& 3.64  \\
        \bottomrule
        \end{tabular}
        }
\vspace{0.5em}
\begin{tablenotes}
 \item[2]* U, B represent UPO, Bicocca respectively.
\end{tablenotes}
\end{threeparttable}
\label{tab:localization-real-fmp1.5}
\end{table}

\textbf{Another Architecture Implementation.} In this paper, we propose two policy architectures and implement them in three specific tasks. Actually, there are many implementations of the two architectures. In this part, we provide another implementation of Policy Architecture 1 in the localization task, which is named \textbf{FMP-1.5}. In this implementation, the memory encoder is of the same architecture as the observation encoder with other designs same as \textbf{FMP-1}. Compared results in three synthetic environments and three real-world environments are shown in Table~\ref{tab:localization-synthetic-fmp1.5} and Table~\ref{tab:localization-real-fmp1.5}. In synthetic environments, \textbf{FMP-1.5} shows comparable average episode success rates to \textbf{FMP-1} with less average trials. In real-world environments, \textbf{FMP-1} shows better performance in Bicocca maps. However, \textbf{FMP-1.5} achieves more than twice that of \textbf{FMP-1} in episode success rate with about the same trial times. Overall, \textbf{FMP-1.5} shows better performance in synthetic maps and UPO map, which demonstrates the advantage of using an embedding representation of $m_t$. But in Bicocca maps with relatively more symmetric structures, which lead to multi-modal predictions, directly leveraging $m_t$ is more effective. This might be due to the fact that a proper embedding representation of $m_t$ might cost more training samples.

\subsection{Case Visualization} 
\begin{figure*}[t]
  \centering
  \includegraphics[width=\linewidth]{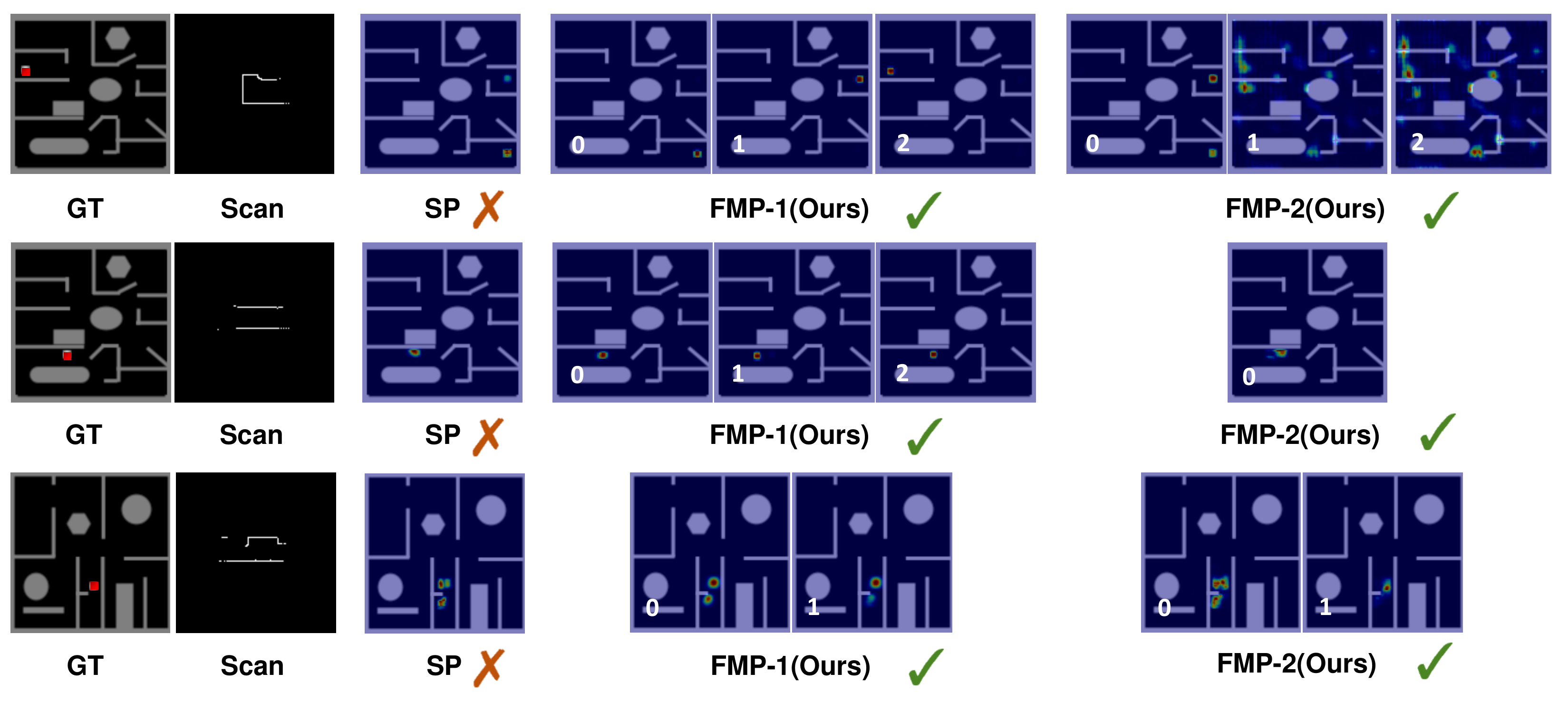}
  \caption{Testing cases of localization task in synthetic environments of three policies. The left two columns show the global map with the ground truth position labeled as a red point, and the scan observation. Other columns show the prediction process and the distributions of three policies. The distribution is reflected by the color, where the value comes larger as the color comes closer to red. \Checkmark means that the policy successfully finds the right position, while \XSolidBrush means a failure. }
  \label{fig:case-localization-more}
\end{figure*}

\begin{figure*}[t]
  \centering
  \includegraphics[width=\linewidth]{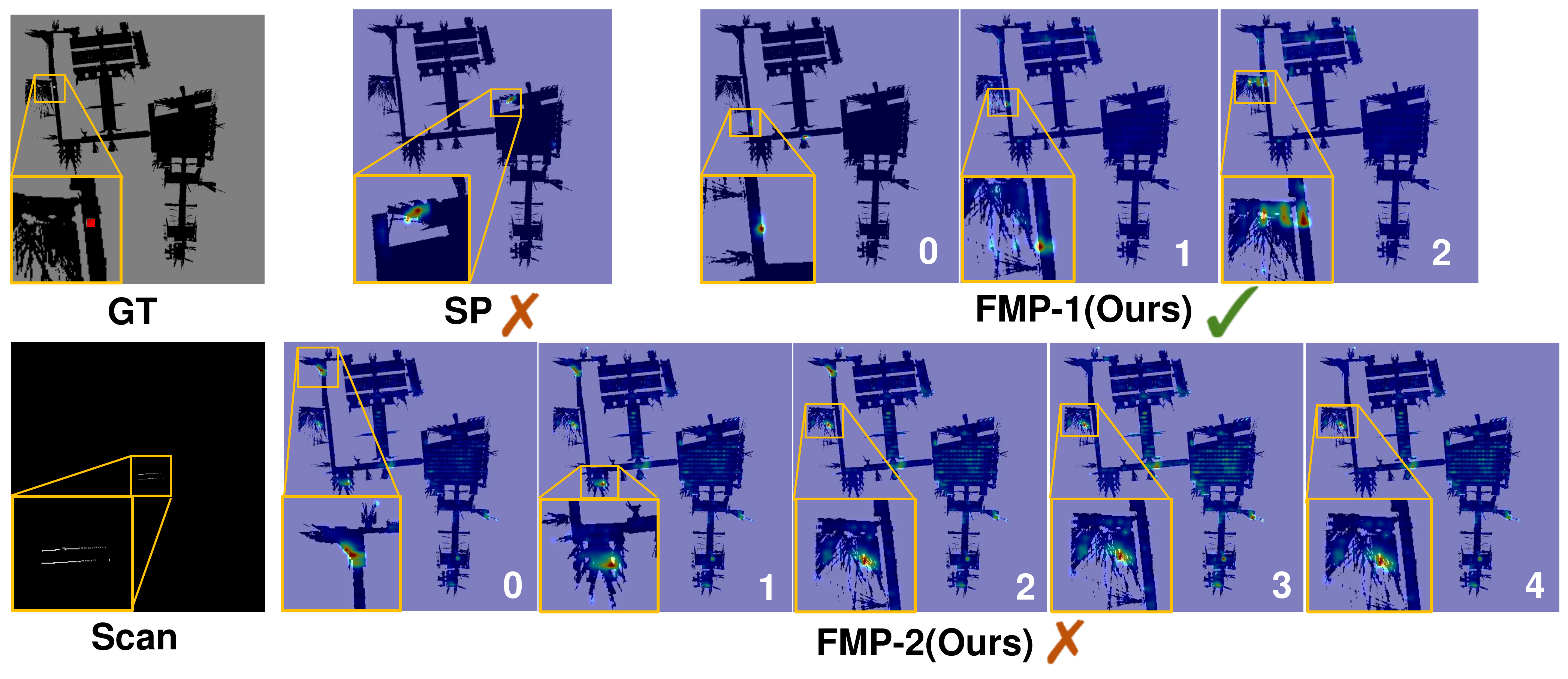}
  \caption{A testing case of localization task in real-world environments of three policies. The left column shows the global map with the ground truth position labeled as a red point, and the scan observation. Other columns show the prediction process and the distributions of three policies. The distribution is reflected by the color, where the value comes larger as the color comes closer to red. \Checkmark means that the policy successfully finds the right position, while \XSolidBrush means a failure. }
  \label{fig:case-localization-real-more}
\end{figure*}

\begin{figure*}[t]
\centering
\subfigure[Normalized probability distributions of all feasible poses the selected pose at all trial steps, and the stars label successful trials.]{
\begin{minipage}[b]{\linewidth}
\includegraphics[width=\linewidth]{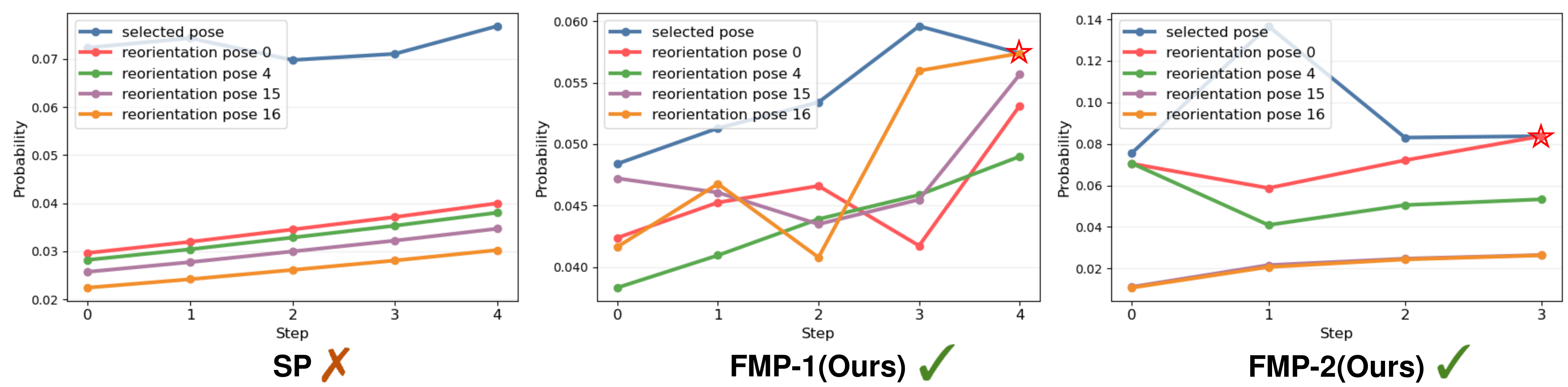}
\end{minipage}
}
\subfigure[Normalized probability distribution changes of the remaining actions at all trial steps of three policies.]{
\begin{minipage}[b]{\linewidth}
\includegraphics[width=\linewidth]{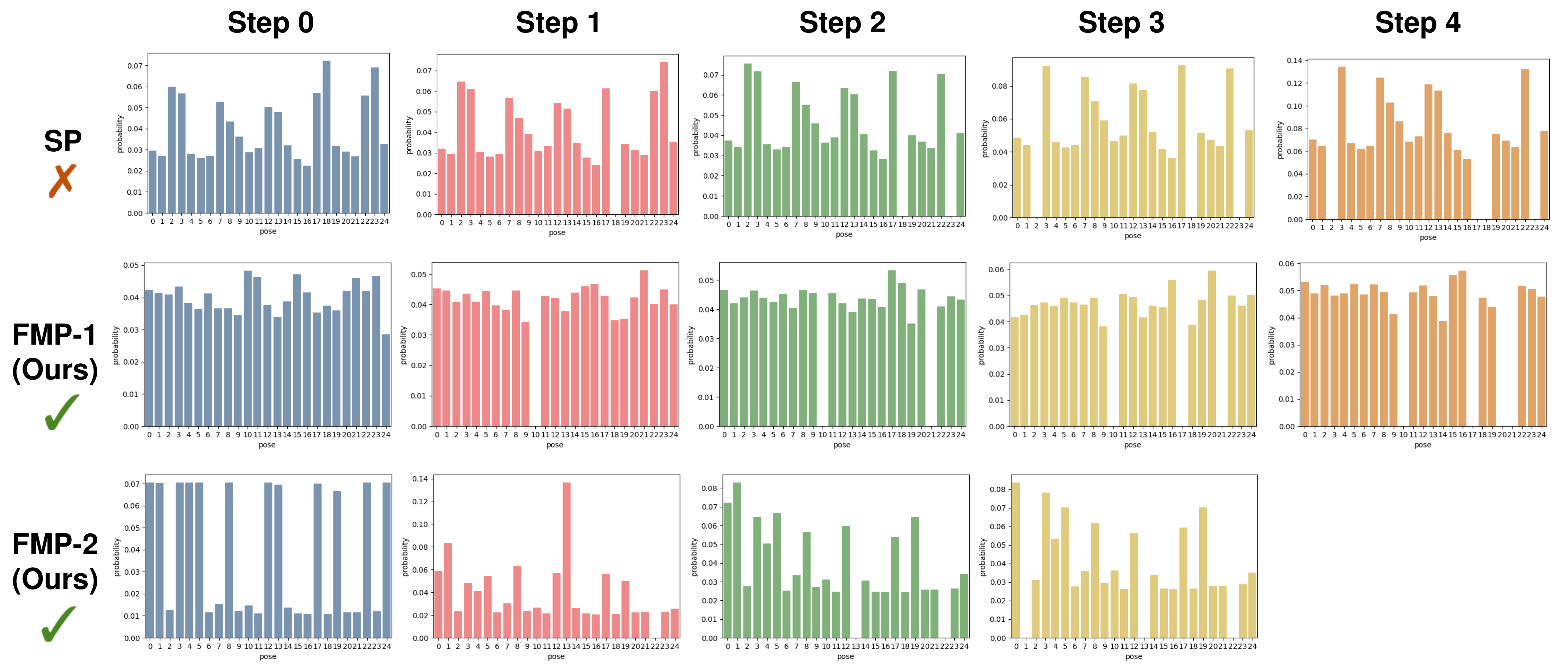}
\end{minipage}
}
\caption{A testing case in reorientation task of three policies. Each case has several feasible poses. \Checkmark means that the policy successfully finds the right position, while \XSolidBrush means a failure. The settings of this case are: conditioner, initial pose: (0.20, -0.29, 0.02, 1.56, -1.51e-05, -3.14), target pose: (-0.05, -0.09, 0.03, 1.04, 1.46, 2.61), gt poses: (0, 4, 15, 16). In this case, the decision sequence of three policies are \textbf{SP}: 18→23→2→17→3, \textbf{FMP-1}: 10→21→17→20→16, \textbf{FMP-2}: 22→13→1→0.}
\label{fig:case-reorientation-more}
\end{figure*}

Fig.~\ref{fig:case-localization-more} and Fig.~\ref{fig:case-localization-real-more} present more testing cases of the localization task in synthetic environments and real-world environments, which further demonstrate the advantage of our failure-aware policies and the different properties of the two policy architectures. Also, Fig.~\ref{fig:case-localization-real-more} is a failure case of \textbf{FMP-2}, which explores with a conservative way in this case. Also, Fig.~\ref{fig:case-reorientation-more} shows the distribution changes at all trial steps of three policies.

\subsection{Detailed Results}
\begin{figure*}[t]
  \centering
  \includegraphics[width=\linewidth]{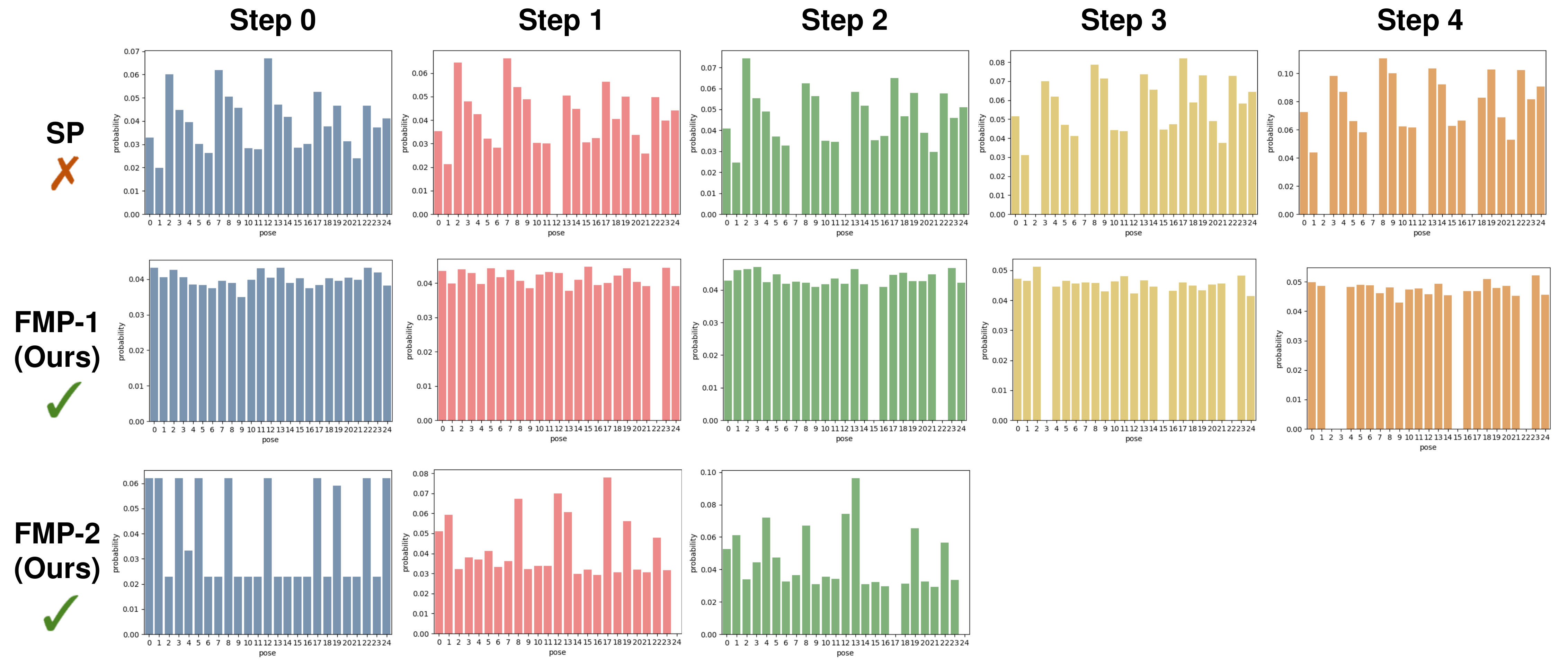}
  \caption{Normalized probability distribution changes of the remaining actions at all trial steps of the case in Fig.~\ref{fig:case-reorientation}. \Checkmark means that the policy successfully find the right pose, while \XSolidBrush means a failure. The decision sequences of three policies are \textbf{SP}: 12→7→2→17→8, \textbf{FMP-1}: 22→15→3→2→23, \textbf{FMP-2}: 24→17→13.}
  \label{fig:case-reorientation-real-detailed}
\end{figure*}

\begin{figure*}[t]
  \centering
  \includegraphics[width=\linewidth]{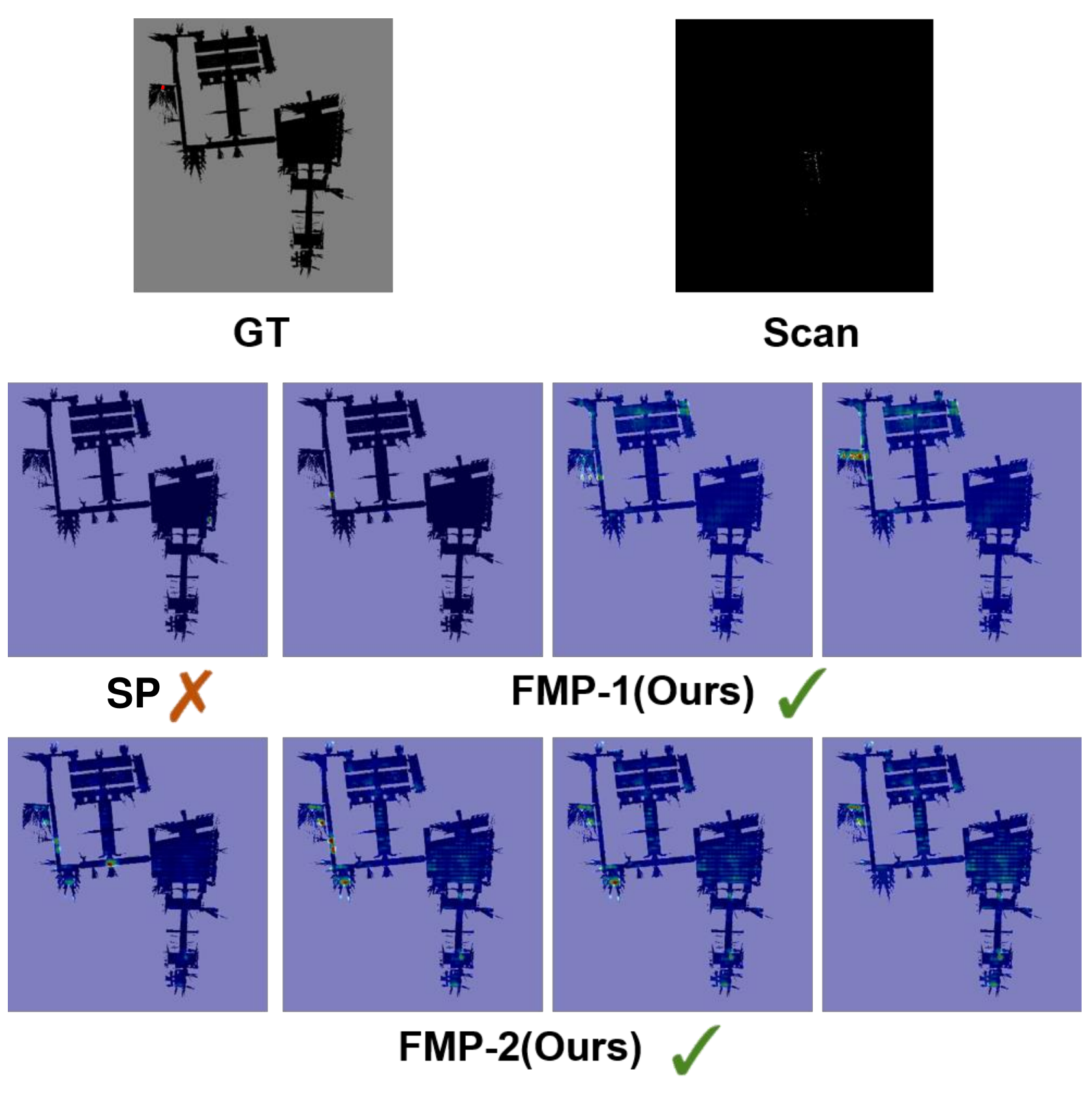}
  \caption{A clear figure of Fig.~\ref{fig:case-localization}.} 
  \label{fig:case-localization-real-detailed}
\end{figure*}

We provide detailed results of Table~\ref{tab:localization-syn} in Table~\ref{tab:localization-synthetic-detail}, which presents the results of all synthetic testing sequences. Also, a clear figure of Fig.~\ref{fig:case-localization} is shown as Fig.~\ref{fig:case-localization-real-detailed}. And Fig.~\ref{fig:case-reorientation-real-detailed} shows the predicted probability distribution changes of all candidate reorientation poses of Fig.~\ref{fig:case-reorientation}.


\end{document}